\documentclass[letterpaper]{article} 
\usepackage{aaai24}  
\usepackage{times}  
\usepackage{helvet}  
\usepackage{courier}  
\usepackage[hyphens]{url}  
\usepackage{graphicx} 
\usepackage{natbib}  
\usepackage{caption} 
\usepackage{algorithm}
\usepackage{algorithmic}

\usepackage{newfloat}
\usepackage{listings}
\DeclareCaptionStyle{ruled}{labelfont=normalfont,labelsep=colon,strut=off} 
\floatstyle{ruled}
\newfloat{listing}{tb}{lst}{}
\floatname{listing}{Listing}
\usepackage{multirow}
\usepackage{array}
\usepackage{amsthm}

\usepackage{mdwmath}
\usepackage{mdwtab}
\usepackage{eqparbox}

\usepackage{amsmath}
\usepackage{amssymb}
\usepackage{amsfonts}
\usepackage{amsmath}
\usepackage{multicol}
\usepackage{bm}
\usepackage{cancel}
\usepackage{subeqnarray}
\usepackage{cases}
\usepackage{mathrsfs}
\usepackage{subfigure}

\usepackage{mdwmath}
\usepackage{mdwtab}
\usepackage{eqparbox}
\usepackage{url}

\newtheorem{theorem}{Theorem}
\newtheorem{proposition}{Proposition}

\newtheorem{myDef}{Definition}

\usepackage{newfloat}
\usepackage{listings}
\usepackage{booktabs}

\title{Towards Inductive Robustness: Distilling and Fostering Wave-induced Resonance in Transductive GCNs Against Graph Adversarial Attacks}
\author{
    Ao Liu\textsuperscript{\rm 1}, Wenshan Li\textsuperscript{\rm 2}\thanks{Corresponding author}, Tao Li\textsuperscript{\rm 1}, Beibei Li\textsuperscript{\rm 1}, Hanyuan Huang\textsuperscript{\rm 1}, Pan Zhou\textsuperscript{\rm 3}
}
\affiliations{

    \textsuperscript{\rm 1}Sichuan University\\
     \textsuperscript{\rm 2}Chengdu University of Information Technology\\
     \textsuperscript{\rm 3}Huazhong University of Science and Technology\\
     \{liuao, huanghanyuan\}@stu.scu.edu.cn, helenali@cuit.edu.cn, \{litao, libeibei\}scu.edu.cn, panzhou@hust.edu.cn

}

\usepackage{bibentry}

\begin{document}

\maketitle

\begin{abstract}
Graph neural networks (GNNs) have recently been shown to be vulnerable to adversarial attacks, where slight perturbations in the graph structure can lead to erroneous predictions.
However, current robust models for defending against such attacks inherit the transductive limitations of graph convolutional networks (GCNs). As a result, they are constrained by fixed structures and do not naturally generalize to unseen nodes.
Here, we discover that transductive GCNs inherently possess a distillable robustness, achieved through a wave-induced resonance process. Based on this, we foster this resonance to facilitate inductive and robust learning.
Specifically, we first prove that the signal formed by GCN-driven message passing (MP) is equivalent to the edge-based Laplacian wave, where, within a wave system, resonance can naturally emerge between the signal and its transmitting medium. This resonance provides inherent resistance to malicious perturbations inflicted on the signal system.
We then prove that merely three MP iterations within GCNs can induce signal resonance between nodes and edges, manifesting as a coupling between nodes and their distillable surrounding local subgraph.
Consequently, we present Graph Resonance-fostering Network (GRN) to foster this resonance via learning node representations from their distilled resonating subgraphs.
By capturing the edge-transmitted signals within this subgraph and integrating them with the node signal, GRN embeds these combined signals into the central node's representation.
This node-wise embedding approach allows for generalization to unseen nodes.
We validate our theoretical findings with experiments, and demonstrate that GRN generalizes robustness to unseen nodes, whilst maintaining state-of-the-art classification accuracy on perturbed graphs.

\end{abstract}

\section{Introduction}

In recent years, graph neural networks (GNNs), through the capabilities afforded by inductive learning, have emerged as the most potent instruments for node classification tasks.
Nevertheless, earlier transductive models, such as graph convolutional networks (GCNs)~\cite{Kpif_2017_ICLR}, have inadvertently introduced vulnerabilities to adversarial attacks within the GNN framework.
It has been observed that perturbed graphs derived from GCNs serving as surrogate models have the potential to compromise the outputs of inductive GNNs when transferred.
In real-world applications, where trust and accuracy are non-negotiable~\cite{chen2021catgcn,nadal2021graph,zhao2022connecting,berberidis2019node,xiao2019knowledge}, such vulnerabilities can significantly jeopardize public trust~\cite{Kreps2020SinceAdv}, distort human decision-making~\cite{Walt2019NMI}, and threaten human well-being~\cite{Samuel2019Sci}. Addressing the vulnerabilities introduced by transductive GCNs into the GNNs' community is of paramount importance.

Distinct from discrete feature data like images or text, graph data comprises a connected set of features through its topological structure. This interconnectedness naturally encourages the adoption of a global input-output mechanism to establish a learning channel from features to labels, a paradigm referred as transductive learning~\cite{Kpif_2017_ICLR,defferrard2016convolutional,bruna2013spectral}, with GCNs epitomizing this approach. This very transductive nature of GCNs offers adversaries an ideal environment for launching attacks~\cite{Liu_2022_TNNLS}. Leveraging this global input-output pattern, given sufficient computation, adversaries can invariably devise perturbations that are both concealment and effective~\cite{sun2022adversarial}.
Given that adversaries exploit vulnerabilities inherent to transductive models to compromise the GNNs' communities, the formulation of a more robust transductive model has ascended as the prevailing defensive approach.

To defense adversarial attacks, early research predominantly sought to fortify GCN's tolerance to perturbations by adversarial training through random edge drops~\cite{DaiLT18ADV_GRA}.
Recently, a shift towards self-supervised training methods has been observed. These techniques sidestep the trap set by adversaries, which bait the model into misclassifying specific inputs. Instead of singularly focusing on enhancing the model's robustness to a given label space, they aim to expand the GCN's overall robustness to potential perturbed graphs. Key representatives of these research endeavors include:
(1) In RGCN~\cite{Zhu2019RGCN}, the Gaussian distributions are employed to replace the node hidden representations across each GCN layer, aiming to mitigate the adversarial modifications' impact.
(2) By introducing a singular value decomposition (SVD) filter before the GCN processing, GCN-SVD~\cite{entezari2020SVD} is designed to discard adversarial edges from the training dataset.
(3) STABLE~\cite{li2022STABLE} introduces enhancements in GCN's forward propagation by incorporating functions that sporadically reinstate edges which were approximately removed.
(4) EGNN~\cite{liu2021EGNN} leverages graph smoothing techniques to confine the permutation setting space, effectively excluding the majority of non-smooth permutations.

However, current research, aiming to improve GCN-based models into a robust transductive variant against attacks, inadvertently carries over transductive-introduced weaknesses~\cite{hamilton_2017_SAGE}. Specifically, these models can't handle unseen nodes and are limited to fixed structures, lacking generalization. This restricts their applicability. If adversaries slightly adjust tactics, defenders must retrain their models for safety. The cause is that GCNs' vulnerabilities are inherent. To enhance their robustness, these vulnerabilities require targeted solutions. Deviating from the context of GCNs could hinder a thorough analysis of attack mechanisms. This, in turn, would obstruct the transition from transductive robust models to inductive ones. Until we harness GCN's inherent robustness for inductive models, we will be stuck in a cycle of constantly refining transductive ones to address vulnerabilities.

In addressing this conundrum, our exploration unveiled an intriguing intrinsic source of robustness within the GCN itself. Without resorting to additional designs, merely deepening the standard 2-layer GCN to a 3-layer structure endows it with an innate (albeit partial) robustness. Importantly, the mechanism underpinning this robustness can be distilled. By purposefully fostering this intrinsic mechanism, it has paved the way for us to architect a robust inductive model. Employing this approach serves a dual purpose: On one hand, it facilitates a precision-oriented confrontation against the perturbations devised specifically by adversaries for transductive structures, ensuring the efficacy of our defensive strategies. On the other hand, it enables us to integrate this robustness mechanism into inductive frameworks, thereby achieving a seamless melding of inductiveness and robustness.

Specifically, we demonstrate that the vibrations of node signals within the GCN-driven message passing (MP) are equivalent to the edge-based waves, formulated by wave equations~\cite{friedman2004wave,shatah1993regularity}. Given this equivalence, it follows that GCNs inherently possess the potential for resonance~\cite{kovalyov1989resonance}, allowing them to harness the natural advantages of waves in defending against perturbations~\cite{blas2022detecting}.
Then, we introduce a mathematical definition for the intensity of such resonance in GCNs. This definition, which outlines the scope and weights of a node's connections to its neighbors, concurrently adheres to four principles: universality, adaptation via MP, node-independence, and topological correlation.
Subsequently, we demonstrated that for 3+ layer GCNs, an invariant mapping exists, translating GCNs' outputs into resonance intensity, manifesting as nodes capturing their surrounding local weighted structure.

Informed by these insights, we introduce the Graph Resonance-fostering Network (GRN) for inductive learning. The core of GRN is that it distills the structure resonating with nodes as local resonance subgraphs.
Then, within this subgraph, GRN fosters the resonance by embedding both the node's signals and the signals transmitted through edges as central node's representation.
This embedding approach is generalizable across graph structures. If the surrounding topology of a node (with unseen ones) can be clearly determined to distill the local resonance subgraph, robust and inductive graph learning is achieved. Our contributions are:

\begin{itemize}
  \item We propose the first inductive and robust GNN.
  \item We prove that a 3-layer GCN inherently possesses an distillable robustness.
  \item We prove the equivalence between GCN-driven signal vibrations and edge-based waves.
\end{itemize}

\section{Preliminaries}
\subsubsection{Notations} We consider connected graphs $\mathcal{G}=(\mathcal{V},\mathcal{E})$ consisting $N=|\mathcal{V}|$ nodes.
Let $\mathbf{A}\in \{0,1\}^{N \times N}$ be the adjacency matrix.
Let generic symbol $\mathbf{L}$ be the Laplacian in its broadest sense.
The feature and one-hot label matrix are $\mathbf{Z} \in \mathbb{R}^{N\times d_0}$ and $\mathbf{Y} \in \mathbb{R}^{N\times d_L}$ respectively.
The edge connected nodes $v_i$ and $v_j$ is written as $(v_i, v_j)$ or $(v_j, v_i)$.
The neighborhood $\mathcal{N}_i$ of a node $v_i$ consists of all nodes $v_j$ for which $(v_i, v_j) \in \mathcal{E}$.
Let $\mathrm{deg}_i$ be the degree of node $v_i$.
The feature vector and one-hot label of node $v_i$ are $\mathbf{z}_i$ and $\mathbf{y}_i$.


\subsubsection{GCN}Under the topology $\mathbf{L}$, with $\mathbf{Z}$ as the input, the output at the $k$-th layer of a GCN is denoted as $\mathcal{M}(\mathbf{Z},k;\mathbf{L})$. The $k$-th parameter matrix of $\mathcal{M}$ is $\mathbf{W}^{(k)} \in \mathbb{R}^{d_k \times d_{k+1}}$. $\mathbf{Z}^{(k)}=[\mathbf{z}_1^{(k)},\ldots,\mathbf{z}_N^{(k)}]$ denotes the features in $k$-th MP.

\subsubsection{Wave Equation}The edge-based wave equation introduces a relationship between a graph-based signal $g = \mathrm{W{\scriptstyle AVE}}(\mathbf{Z},k;\mathbf{L})$ and its topological structure. Let $c$ be a constant, it is defined as $\frac{\partial^2 g}{\partial k^2}=-\mathbf{L}^{k}g \cdot c$~\cite{friedman2004wave}. Herein $g$ can be instantiated as any discernible signal.

\section{3-layer GCN Possesses Adversarial Robustness via Wave-induced Resonance}\label{sec_res}

\subsubsection{Equivalence of GCN-driven MP and Wave Equation}
Here we demonstrate that the signal vibrations driven by GCNs, are equivalent to waves on graph topologies and can be characterized by nonlinear wave equations.

\begin{theorem}\label{thm_1}
Let $\mathcal{M}(\mathbf{Z},k;\mathbf{L})$ and $\mathrm{W{\scriptstyle AVE}}(\mathbf{Z},k;\mathbf{L})$ denote the signals of the $k$-th MP and $k$-th wave respectively, under the topological structure represented by $\mathbf{L}$. It is established that for the given $k$ and $\mathcal{M}(\cdot)$, there exists $\mathrm{W{\scriptstyle AVE}}(\cdot)$ satisfies $\mathcal{M}(\mathbf{Z},k;\mathbf{L}) = \mathrm{W{\scriptstyle AVE}}(\mathbf{Z},k;\mathbf{L}), \forall \mathbf{L} \in \widehat{\mathbf{L}}$,
where $\widehat{\mathbf{L}}$ are the Laplacian matrices of all attribute graphs.
\end{theorem}

This study draws an analogy between the node signals in GCN-driven MP and waves, considering edges as the transmission medium. Research indicates that in systems producing waves, resonance can arise between waves and their medium~\cite{ahmad2023resonance,bykov2019coupled}.
Building on this understanding, we can reaffirm our empirical observations about the GCN training pattern: in non-adversarial contexts, GCNs converge to the most natural, or congruent with ground truth, signal MP paradigm during training. Under this premise, the messages transmitted by nodes and edges in the graph manifest a natural coupling state, maintaining a benign mapping relationship $\mathcal{M}: \mathcal{G} \to \mathbf{Y}$. The key to optimizing $\mathcal{M}$ is the resonance between node signals $\mathbf{Z}^{(\ell)}$ and edge signals $\mathbf{E}^{(\ell)}$.
In adversarial situations, adversaries manipulate node signals by rewiring edges, which inadvertently induces unnatural, i.e., noncongruent with ground truth, MP patterns. Under this scheme, the benign resonance is disrupted, resulting in a malignant mapping relationship $\mathcal{M}: \mathcal{G}' \to \mathbf{Y}'$, where $\mathcal{G}'$ and $\mathbf{Y}'$ is the perturbed graph and label, respectively.

\subsubsection{Mathematical Definition of Resonance in GCN}
Maintaining benign resonance becomes an intuitive defensive mechanism as it intrinsically resists unnatural perturbations. To actualize control over this resonance, thereby purposefully fostering resonance within the GCN, we subsequently delineate a detailed definition of this resonance. Thus, this definition should comply with the following conditions:
1) Every node within a graph should possess a computable resonance intensity,
2) the resonance intensity of all nodes should evolve in accordance with MP,
3) each node should maintain an independent resonance intensity, and
4) the stronger a node's connection to its surrounding topology, the greater its perceived resonance intensity.

To devise a methodology compliant with the desired conditions, we consider node $v_i$ and utilize its latent representation~\cite{Maciej2015sumfeature} $\bar{{z}}_{i}^{(k)}=\sum_j \mathbf{z}^{(k)}_{i,j}$ to quantify the intensity of the node signals.
Furthermore, we use $T_i$, the count of edges among nodes in $\mathcal{N}_i$, to measure the connectivity strength specific to the edges at the given nodes.
Then, we use the total number $p_i=\sum_{j}\mathbf{A}^2_{i,j}$ of walks of length 2 originating from $v_i$ to any node in $\mathcal{G}$, to quantity the magnitude of connectivity density that $v_i$ exhibits in the structure.

Accordingly, we propose the following definition to quantify the resonance intensity at node $v_i$:
\begin{myDef}\label{def_RI}
  The resonance intensity of $v_i$ on $k$-th MP is
  \begin{equation}\label{eq_define_resonance}
R(v_i;k) \overset{\text{def.}}{=}  \bar{{z}}_{i}^{(k)} T_i +  2 p_i + 8 \mathrm{deg}_i.
  \end{equation}
\end{myDef}

The unique of defining resonance intensity can be articulated as follows: it not only allows for an interpretable quantification of the resonance on different nodes, but it is also directly observable within MP. This implies that under such a definition, the resonance intensity of any node at any given MP epoch on a graph can be independently calculated, obviating the need for the GCN computational paradigm.

\subsubsection{Resonance arises in 3rd MP}
Definition~\ref{def_RI} facilitates the quantification of resonance for any signal function on any graph, irrespective of whether or not it is driven by GCN. Nonetheless, an intriguing finding has been proven: the wave system constructed by GCN inherently and involuntarily arises resonance, which is outlined in the theorem:
\begin{theorem}\label{thm_2}
Let $\bar{{z}}_{i}^{(k)}$ be the latent signal formed by GCN-driven MP, we have:
\begin{equation}\label{eq_thm_2}
   R(v_i;k) \propto 64 \bar{{z}}_{i}^{(k+3)} - 32.
\end{equation}
\end{theorem}

Theorem~\ref{thm_2} unveils an intriguing phenomenon: for $k\geq3$, there subsists an invariant mapping, which transposes the GCN-driven signal into a resonance intensity that bears no correlation with the GCN paradigm. Given that Definition~\ref{def_RI} has established the resonance intensity as a measure of the coupling strength between nodes and structure within the graph, we can thus characterize it as the degree of coupling. Consequently, it can be asserted that prior to 3rd MP iteration, the GCN appears to have yet to delve into the coupling paradigm between nodes and structure within the graph. However, subsequent to the 3rd MP, due to the persistent presence of the invariant mapping, it can be construed that the GCN has fortuitously assimilated the coupling paradigm within the graph during the 3rd MP, and perpetuates this paradigm into subsequent MPs.

\subsubsection{Vast Perturbation Search Space of 3-layer GCN}

In light of the current absence of an effective method for quantifying the combined adversarial robustness of a specific graph and a GCN learning from said graph, we propose an intuitive approach. For a graph $\mathcal{G}$, comprised of $|\mathcal{V}|$ edges and represented by the adjacency matrix $\mathbf{A}$, and a GCN $\mathcal{M}$ with $K$ layers, where the perturbation budget is denoted as $r$, the number of matrix multiplication-based forward propagations required by the attack model can be construed as the highest attack cost. In this context, the number of subgraphs is independent of node features, hence we employ $\mathbf{A}$ as the independent variable for the attack cost function, denoted as $\mathrm{Cost}(\mathbf{A},r,K)$. We then present the following theorem:
\begin{theorem}\label{thm_cost}
For any specified graph with a node set $\mathcal{V}$ and an adjacency matrix $\mathbf{A}$, in conjunction with a $K$-layer GCN, and a maximum perturbation $r$, the following holds:
{\small
   \begin{align}\label{eq_cost}
        \mathrm{Cost}(\mathbf{A},r,K) \leq
        \begin{cases}
          C(|\mathcal{V}|,r), & \mbox{if } K < 3 \\
          (K-1) C\left(\frac{|\mathcal{V}|^{K-1}}{2},r\right), & \mbox{otherwise},
        \end{cases}
   \end{align}
}
where $C(\cdot,\cdot)$ denotes the number of combinations.
\end{theorem}

It's revealed that adversaries face the same computational cost for matrix multiplication-based forward propagations when $K$ = 1 or 2. However, for $K \geq 3$, the cost dramatically increases, largely due to $C(\frac{|\mathcal{V}|^{K-1}}{2},r)$. As an example, with the Cora dataset ($|\mathcal{V}|=5429$) and a 1\% perturbation rate ($r=54$), the cost for $K$ = 3 becomes exponentially larger. Thus, attacking a 3-layer GCN presents a vast search space for adversaries. This insight extends Theorem~\ref{thm_2}'s real-world applications and our previous findings: a 3-layer GCN can naturally create resonance robustness. With our defined resonance, we can further boost this robustness proposefully.

\section{Graph Resonance-fostering Network}

\subsubsection{Principle Overview}
We employ GRN to enhance the resonance of the GCN. The underlying concept of the GRN is articulated as follows. Definition~\ref{def_RI} exhibits that for a node $v_i$, there exists a local graph structure that resonates, known as the local resonance subgraph (LRS) for node $v_i$, denoted as $G_i=(\mathcal{V}_{G_i},\mathcal{E}_{G_i})$, used to represent the maximal subgraph structure that node $v_i$ can capture. During end-to-end training, both the node signals $\mathbf{Z}^{(\ell)}$, and the signals transmitted through edges $\mathbf{E}^{(\ell)}$ concurrently vibrate within the LRS. Consequently, for $v_i$, if a learnable parameter $\mathbf{W}^{(\ell)}$ capable of jointly embedding MP's result $\mathbf{A}_{G_i}\mathbf{Z}^{(\ell)}_{G_i}$, and $\mathbf{E}^{(\ell)}_{G_i}$, into $v_i$'s output representation, this aggregation intentionally accomplishes a learnable resonance, generating a local-level embedding. This identical aggregate pattern is applied across all nodes to facilitate a mapping, thereby achieving a global-level forward propagation within the GRN.

In summary, a single forward propagation of the GRN is:
\begin{equation}\label{eq_RFL_FP}
    \mathbf{z}^{(\ell+1)}_i=\sigma( \mathrm{\scriptstyle MEAN} ( \mathrm{\scriptstyle CONCAT} ( \mathbf{A}_{G_i} \mathbf{Z}^{(\ell)}_{G_i} ,\mathbf{E}^{(\ell)}_{G_i} )  \mathbf{W}^{(\ell)} ) ).
\end{equation}

Next, we provide explicit definitions for $G_i$ and $\mathbf{E}^{\ell}_{G_i}$.

\subsubsection{$G_i$: Local Resonance Subgraph}

\begin{figure}[htb]
\centering
\includegraphics[width=0.46\textwidth]{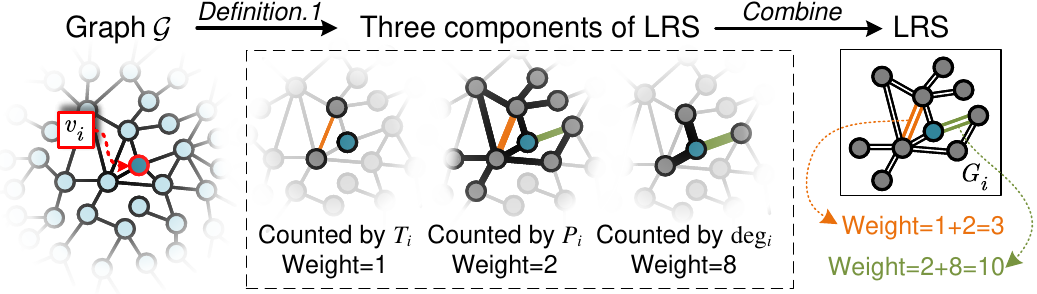}
\caption{An illustration of local resonance subgraph.}
\label{fig_LRS}
\end{figure}

As per Definition~\ref{def_RI}, LRS comprises three components: 1) edges formed amongst all first-order neighbors, as counted by $T_i$, with these edges' weights equal 1; 2) edges formed between it and all 2nd- (inevitably includes 1st-) order neighbors, as counted by $p_i$, with these edges' weights equal 2; 3) edges between it and all first-order neighbors, as counted by $\mathrm{deg}_i$, with these edges' weights equal 8. Consequently, the LRS can be viewed as a weighted graph, in which the weights of edges serve as attention for the joint combination of $\mathbf{Z}^{(\ell)}$ and $\mathbf{E}^{(\ell)}$. An illustrative example of the LRS is presented in Figure~\ref{fig_LRS}.

\subsubsection{$\mathbf{E}^{(\ell)}_{G_i}$: Edge-transmitted Signals}\label{sec_ETS}

In the MP driven by adjacency matrices, only signals at the nodes are observable, while the signals transmitted across each edge remain imperceptible. To ascertain the quantified signals on every specific edge within $G_i$, we first obtain the global edge-transmitted signals $\mathbf{E}^{\ell}_{\mathcal{G}}$. Then, $\mathbf{E}^{\ell}_{G_i}$ is subsequently derived through a sampling procedure on $\mathbf{E}^{\ell}_{\mathcal{G}}$ using the edge indices within $G_i$.

\begin{figure}[htb]
\centering
\includegraphics[width=0.49\textwidth]{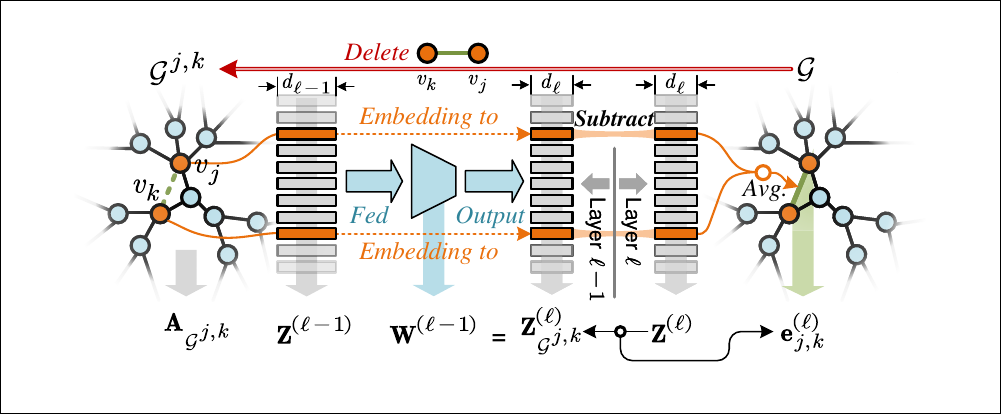}
\caption{Schematic diagram illustrating the computation of the edge-transmitted signal between nodes $v_j$ and $v_k$.}
\label{fig_ETS}
\end{figure}

Specifically, within $\mathbf{E}^{\ell}_{\mathcal{G}}$, the edge-transmitted signals on $(v_j,v_k)$ are denoted as $\mathbf{e}^{\ell}_{j,k}$.
For $\ell>0$, $\mathbf{e}^{\ell}_{j,k}$ is defined via a sequential procedure:
1) The edge $(v_j,v_k)$ in $\mathcal{G}$ is deleted, producing a new graph $\mathcal{G}^{j,k}$ with its adjacency matrix $\mathbf{A}_{\mathcal{G}^{j,k}}$.
2) A new forward propagation is executed in the same layer on $\mathcal{G}^{j,k}$, obtaining a feature matrix $\mathbf{Z}^{(\ell)}_{\mathcal{G}^{j,k}}$. This matrix does not contain any messages transmitted through the edge $(v_j,v_k)$. Consequently,
\begin{equation}\label{eq_Z_G_jk}
 \mathbf{Z}^{(\ell)}_{\mathcal{G}^{j,k}}=\mathbf{A}_{\mathcal{G}^{j,k}}\mathbf{Z}^{(\ell-1)}\mathbf{W}^{(\ell-1)}.
\end{equation}
The feature of node $v_j$ in $\mathbf{Z}^{(\ell)}_{\mathcal{G}^{j,k}}$ denoted as $\mathbf{z}^{(\ell)}_{\mathcal{G}^{j,k},j}$, is obtained.
3) In $\mathcal{G}^{j,k}$, there is no edge between the nodes $(v_j,v_k)$. Hence, the feature transmitted from node $v_k$ to $v_j$ (i.e., $\mathbf{e}^{(\ell)}_{j,k}$) is calculated by subtracting the feature obtained through the re-propagation on $\mathcal{G}^{j,k}$ (i.e., $\mathbf{z}^{(\ell)}_{G_i^{j,k},j}$) from the original feature (i.e., $\mathbf{z}^{(\ell)}$). Similarly, the signal transmitted through the pair $(v_j, v_k)$ could be interpreted as the average of the mutually transmitted signals.
At $\ell=0$, since MP has not been initiated, $\mathbf{e}^{(\ell)}_{j,k}$ would ideally be 0. For end-to-end training, it is defined as a random infinitesimal value.
In conclusion, $\mathbf{E}^{(\ell)}_{G_i}$ is determined as
{\small
\begin{multline}\label{eq_ETS}
  \mathbf{E}^{(\ell)}_{G_i} = \mathrm{\scriptstyle CONCAT}\left( \left\{\mathbf{e}^{(\ell)}_{j,k}: v_j,v_k \in G_i \right\} \right) , \\
  \text{ s.t. } \mathbf{e}^{(\ell)}_{j,k}= \left\{ \begin{array}{l}
                                    \mathbf{z}^{(\ell)} - \frac{\mathbf{z}^{(\ell)}_{\mathcal{G}^{j,k},j}+\mathbf{z}^{(\ell)}_{\mathcal{G}^{j,k},k}}{2}, \mbox{\emph{if}} \ \ell > 0 \\
                                  \epsilon, \quad \text{where } \epsilon \sim U(0, 1\times10^{-7}) \ ,\mbox{\emph{if}} \ \ell = 0
                                  \end{array}\right..
\end{multline}
}
Figure~\ref{fig_ETS} illustrates the computation of $\mathbf{e}^{(\ell)}_{j,k}$.

\subsubsection{Simplifying the Computational Overhead of $\mathbf{E}^{(\ell)}_{G_i}$}
Equation~\eqref{eq_Z_G_jk} explicates the method of re-propagation on $\mathcal{G}_{j,k}$. Given that there are $|\mathcal{V}|$ ways to choose $(v_j, v_k)$, it necessitates the computation of $|\mathcal{V}|$ matrix multiplications (where $\mathbf{Z}_{\mathbf{W}}^{(\ell-1)} = \mathbf{Z}^{(\ell-1)}\mathbf{W}^{(\ell-1)}$ remains the same for all $(v_j, v_k)$ selections and can be considered a constant matrix), thereby constituting the primary computational cost of $\mathbf{E}^{(\ell)}_{G_i}$. Here, we provide a computational method equivalent to Equation~\eqref{eq_Z_G_jk}, reducing the $|\mathcal{V}|$ times to once. 
\begin{proposition}\label{pro_E_overhead}
  By indexing and rearranging $\mathbf{Z}_{\mathbf{W}}^{(\ell-1)}$ by rows $j$ and $k$ to obtain a matrix $Q(\mathbf{Z}_{\mathbf{W}}^{(\ell-1)};j,k) \in \mathbb{R}^{N \times d_{\ell-1}}$,
  \begin{equation}\label{eq_E_overhead}
    \mathbf{Z}^{(\ell)}_{\mathcal{G}^{j,k}}=\mathbf{A}\mathbf{Z}_{\mathbf{W}}^{(\ell-1)}-Q\left(\mathbf{Z}_{\mathbf{W}}^{(\ell-1)};j,k\right).
  \end{equation}
\end{proposition}
  Evidently, a single matrix multiplication, i.e., $\mathbf{A}_{G}\mathbf{Z}_{\mathbf{W}}^{(\ell-1)}$, is sufficient to iterate over all $(v_j, v_k)$ and yield the results.

\begin{figure}[htb]
\centering
\includegraphics[width=0.49\textwidth]{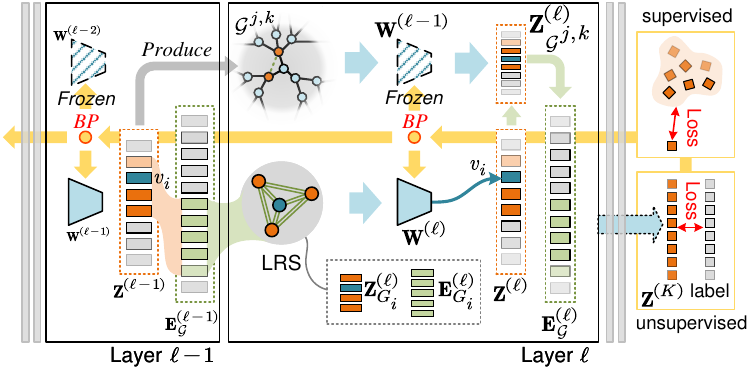}
\caption{The general workflow of GRN.}
\label{fig_RFL}
\end{figure}

\subsubsection{Learning the Parameters} Each layer of GRN only contains trainable parameters $\mathbf{W}^{(\ell)}$, and each has a distinct output representation $\mathbf{Z}^{(\ell)}$. Thus, in accordance with the requirements of the downstream task, GRN can accommodate either supervised or unsupervised loss functions, thereby tuning their weight matrices. Specifically, we denote the discrepancy function as $D(\cdot,\cdot)$. In semi-supervised scenarios, the loss function is $J_s(\mathbf{z}^{(K)}_i)=D(\mathbf{z}^{(K)}_i, \mathbf{y}_i)$; in unsupervised scenarios~\cite{muller2023graph}, $J_u(\mathbf{z}^{(K)}_i)=D(\mathbf{z}^{(K)}_i, \{\mathbf{y}_j:v_j \in \mathcal{N}_i\})$. Depending on the downstream applications, $D(\cdot,\cdot)$ can take various forms, such as cross-entropy, etc. The general workflow of GRN is illustrated in Figure~\ref{fig_RFL}.


\section{Experiments}

\subsubsection{Datasets} Our findings are evaluated on five real-world datasets widely used for studying graph adversarial attacks~\cite{Sun_2020_WWW,Liu_2022_TNNLS,Zhu2019RGCN,entezari2020SVD,li2022STABLE}, including {Cora}, {Citeseer}, {Polblogs}, and {Pubmed}.

\subsubsection{Baselines}
\underline{\textit{Comparison defending models.}} We compare GRN with other defending models including:
        1) {RGCN} which leverages the Gaussian distributions for node representations to amortize the effects of adversarial attacks,
        2) {GNN-SVD} which is applied to a low-rank approximation of the adjacency matrix obtained by truncated SVD,
        3) {Pro-GNN}~\cite{jin2020Pr0GNN} which can learn a robust GNN by the intrinsic properties of nodes,
        4) {Jaccard}~\cite{wu2019jaccard} which defends attacks based on the measured Jaccard similarity score,
        5) {EGNN}~\cite{liu2021EGNN} which filters out perturbations by $l_1$- and $l_2$-based graph smoothing.
\underline{\textit{Attack methods.}} The experiments are designed under the following attack strategies:
        1) {Metattack}~\cite{Zuger2018METTACK}, a meta-learning based attack,
        2) {CLGA}~\cite{zhang2019CLGA}, an unsupervised attack,
        3) {RL-S2V}~\cite{DaiLT18ADV_GRA}, a reinforcement learning based attack.

\subsubsection{Pinpointing the Layer of Resonance}

In Theorem~\ref{thm_2}, we establish an equivalence relation between the $k$-th and the $k+3$-th layer's output latent representations, as derived from Equations~\eqref{eq_define_resonance} and~\eqref{eq_thm_2}. This elucidates that when $k=0$, the $3$-th layer involuntarily captures local structures, thereby inducing resonance. To facilitate experimental variable control, we first demonstrate the equivalence relation under varying ``gap layer numbers" (denoted as $k_{gap}$). If the equivalence between Equations~\eqref{eq_define_resonance} and~\eqref{eq_thm_2} only holds when $k_{gap}\geq 3$,  it substantiates the validity of Theorem~\ref{thm_2}. Specifically, we first train a 5-layer GCN, then obtain the resonance intensity denoted as $R_{def}(k) = \bar{{z}}_{i}^{(k)} T_i +  2 p_i + 8 \mathrm{deg}_i$, and the actual observed signal denoted as $R_{real}(k+k_{gap}) = 64 \bar{{z}}_{i}^{(k+k_{gap})} - 32$, for each epoch.
Given these observational variables, we delineate their transformations over the learning process using lists $\{ R_{def}(k)\}$ and $\{ R_{real}(k+k_{gap})\}$ respectively. Each list chronicles its corresponding variable's fluctuations across all epochs.
Subsequently, we standardize (using the standardize function $\mathrm{{\scriptstyle STD}}(\cdot)$) the sequences under different $k_{gap}$ and calculate the absolute difference to obtain a difference sequence:
\begin{equation}\label{eq_exp_STD}\small
  \mathbf{d}_{k,k_{gap}} = |\mathrm{\scriptstyle STD}(\{ R_{def}(k)\}) - \mathrm{\scriptstyle STD}(\{ R_{real}(k+k_{gap})\})|.
\end{equation}

\begin{figure}[htb]
\centering
\includegraphics[width=0.48\textwidth]{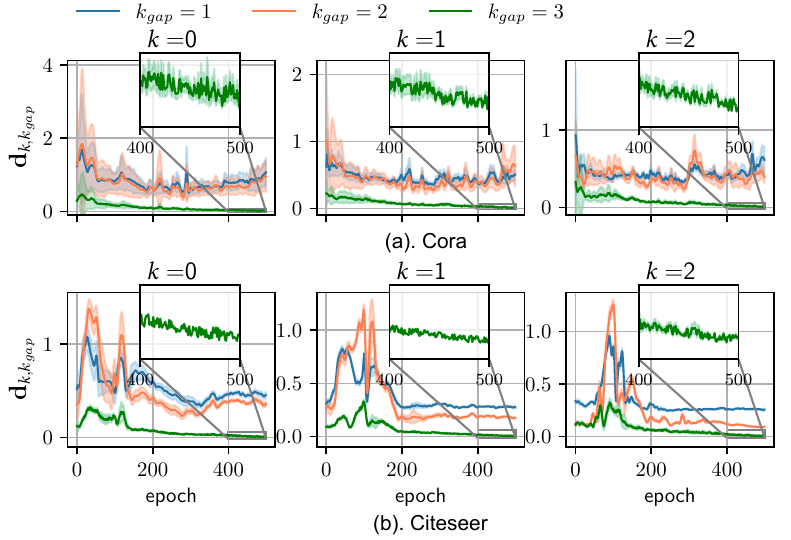}
\caption{Values of $\mathbf{d}_{k,k_{gap}}$ under different $k$ and $k_{gap}$ settings. The $\mathbf{d}_{k,k_{gap}}$ encapsulates the resonance situation between the $k$-th layer and the $k+k_{gap}$-th layer. A smaller value indicates a stronger resonance.}
\label{fig_exp_thm2_main}
\end{figure}

The parameter $\mathbf{d}_{k,k_{gap}}$, serving as an indicator variable, accurately encapsulates the discrepancy between $R_{def}(k)$ and $R_{real}(k+k_{gap})$.
The experimental results are illustrated in Figure~\ref{fig_exp_thm2_main}. Owing to the large number of nodes, we display the mean value (central line) and standard deviation (shadow areas) of all nodes. As epochs progress, $\mathbf{d}_{0,3}$ gradually converges to zero. After the initial several epochs, it significantly diverges from $\mathbf{d}_{0,1}$ and $\mathbf{d}_{0,2}$. This validates the intriguing phenomenon mentioned in Theorem~\ref{thm_2}: a correlation has been established between the signal at the 3-th layer and the graph's initial signal and structure.
Subsequent experimental results echo the aforementioned findings, thereby affirming the correctness of Theorem~2 when $k>0$.

\begin{figure*}[htbp]
    \centering

    \subfigure[Visualizations of embedding of $\mathcal{G}$, $\mathcal{G}_{LRS}$ and $\mathcal{G}_{random}$.]{
        \includegraphics[width=0.44\textwidth]{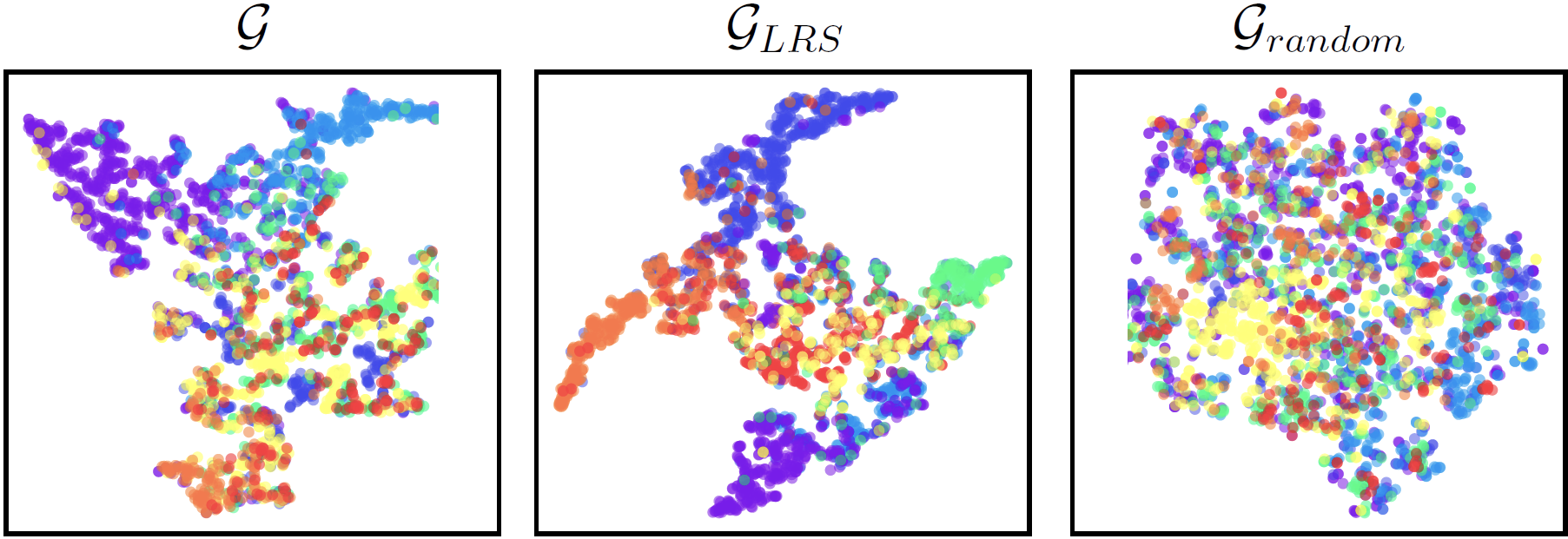}
        \label{fig_LRS_nonADV}
    }
    \subfigure[Strength distribution of $\mathcal{G}$, $\mathcal{G}_{LRS}$ and $\mathcal{G}_{random}$.]{
        \includegraphics[width=0.40\textwidth]{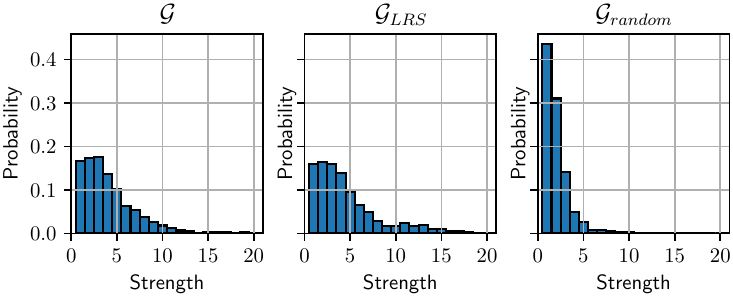}
        \label{fig_LRS_strength}
    }

    \subfigure[Classification accuracy of $\mathcal{G}$, $\mathcal{G}_{LRS}$ and $\mathcal{G}_{random}$ using $\mathcal{M}_{\mathcal{G}}$ under different perturbation rates.]{
        \centering
        \includegraphics[width=0.90\textwidth]{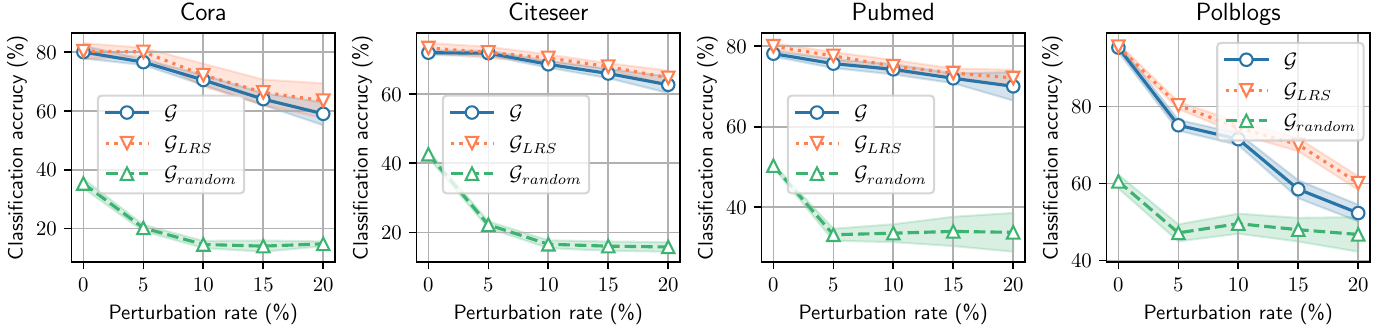}
        \label{fig_LRS_ADV}
    }

    \caption{Experimental results of LRS-constructed graph $\mathcal{G}_{LRS}$ in relation to $\mathcal{G}$}
    \label{fig:my_label}
\end{figure*}

\subsubsection{Attack Success Rate Cliff-like declines on 3-layer GCN}

\begin{figure}[htb]
\centering
\includegraphics[width=0.47\textwidth]{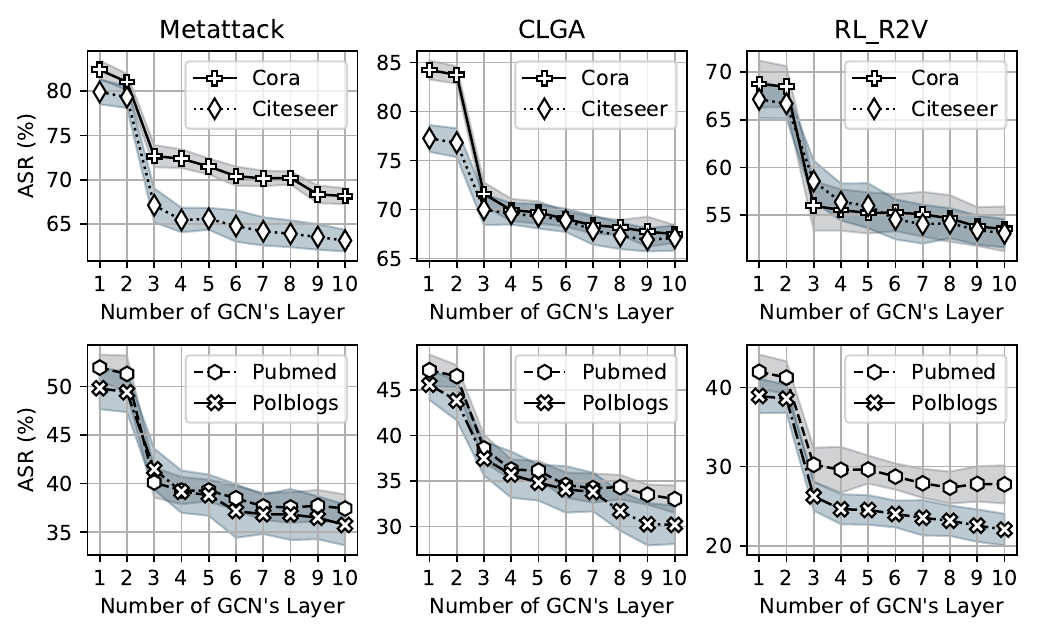}
\caption{GCN's layer count and ASR relation.}
\label{fig_exp_layer_inc}
\end{figure}

 Intuitively, the complexity of an attack tends to increase with the number of GCNs' layer. Observing the pattern of attack success rate (ASR) declines as the number of GCN layers increases helps validate our claim that the 3-layer GCN, derived from resonance, can significantly enhance robustness. Specifically, we start by initializing 10 GCNs with the number of layers ranging from 1 to 10. Next, we conduct experiments on 4 datasets using 3 typical attacks, setting the perturbation rate uniformly at 20\%. We then train a surrogate model for each GCN separately, placing perturbations in the dataset, and clearing these perturbations after each attack. We repeat each attack five times and report the average ASR accuracy (depicted by the lines) and variation range (represented by the shaded areas). The results, as shown in Figure~\ref{fig_exp_layer_inc}, clearly reveal a steep drop in ASR at the 3-layer GCN. However, further layer addition seems unable to significantly reduce the ASR, as the additional layers maintain the same resonance pattern as the 3-layer GCN to achieve adversarial robustness. These findings articulate the concept of distilling the resonance from GCNs and fostering this resonance to design a inductive approach.

\subsubsection{How Robust of LRS-constructed Graphs}

Derived from a transductive model, the LRS captures distinct resonance regions and transforms a localized, unweighted graph (also perceivable as a graph with unitary weights) into a weighted one. The implementation of the LRS within the GRN enables the demarcation of a learnable resonance scope for inductive learning. Consequently, it becomes essential to validate the efficacy of the LRS through its embedding precision within the transductive model.

We initiate by presenting results obtained under non-adversarial conditions. We sum the LRS of all nodes in $\mathcal{G}$ and apply min-max scaling to all weights, thus creating a global weighted graph $\mathcal{G}_{LRS}$. Then, we train a standard 2-layer GCN $\mathcal{M}_{\mathcal{G}}$ on Cora dataset (denoted as $\mathcal{G}$) and visualize its well-trained representations. Then, we feed $\mathcal{G}_{LRS}$ into $\mathcal{M}_{\mathcal{G}}$ to generate its visualization. Lastly, for comparison, we create a random-weighted graph $\mathcal{G}_{random}$ whose edge distribution is the same as $\mathcal{G}$, and input it into $\mathcal{M}_{\mathcal{G}}$ to get the corresponding visualization. As Figure~\ref{fig_LRS_nonADV} shows, under identical weights, the representations of different categories in $\mathcal{G}_{LRS}$ are tighter than those in $\mathcal{G}$. This suggests that introducing LRS brings beneficial global weights, which enhance the model's performance in non-adversarial scenarios.

Then, we explored the similarity between $\mathcal{G}$ and $\mathcal{G}_{LRS}$. Using strength distribution, akin to degree distribution for unweighted graphs~\cite{DanielAS2018ADV_GRA}, we found the two weighted graphs are notably similar. Figure~\ref{fig_LRS_strength} confirms this, showing a stark difference from $\mathcal{G}_{random}$. Therefore, $\mathcal{G}_{LRS}$ maintains the traits of $\mathcal{G}$.

We next assessed the classification accuracy of $\mathcal{G}_{LRS}$ under adversarial attacks using varying Metattack perturbation rates $p_r = \frac{r}{|\mathcal{E}|}$. Figure~\ref{fig_LRS_ADV} shows that as $p_r$ increases, $\mathcal{G}_{LRS}$'s accuracy consistently edges out $\mathcal{G}$. This suggests that the LRS introduces a resonance in the transductive model, marginally boosting its adversarial robustness.

\subsubsection{Generalizable Robustness of GRN}

\begin{table*}[htbp]
  \centering \small

    \begin{tabular}{c | c | ccccc | ccc}
    \toprule
        \multirow{2}[0]{*}{{Dataset}} & \multirow{2}[0]{*}{{$p_r$ (\%)}} & \multicolumn{5}{c }{Defense Baselines} & \multicolumn{3}{c }{GRN}  \\
    \cmidrule(lr){3-7} \cmidrule(lr){8-10}
        &     & RGCN & GNN-SVD & Pro-GNN & Jaccard & EGNN & $s_r$=20\% & $s_r$=40\% & $s_r$=60\% \\
  \cmidrule(lr){1-2}\cmidrule(lr){3-7} \cmidrule(lr){8-10}
    \multirow{4}[0]{*}{{Cora}} & 0   &  83.49\scriptsize{±0.57} & 81.14\scriptsize{±0.79} & 85.01\scriptsize{±0.40} & 81.74\scriptsize{±0.36} & 85.00\scriptsize{±0.40} & 83.74\scriptsize{±1.68} & 86.79\scriptsize{±2.27} & \textbf{87.75\scriptsize{±0.93}} \\
        & 5   &  77.20\scriptsize{±0.47} & 78.29\scriptsize{±0.63} & 80.10\scriptsize{±0.22} & 80.56\scriptsize{±1.30} & 82.24\scriptsize{±0.49} & 81.48\scriptsize{±0.83} & 86.04\scriptsize{±3.15} & \textbf{86.24\scriptsize{±1.54}}\\
        & 10  &  72.65\scriptsize{±0.40} & 70.81\scriptsize{±1.77} & 74.45\scriptsize{±0.28} & 75.07\scriptsize{±1.28} & 76.38\scriptsize{±0.35} & 79.51\scriptsize{±1.62} & 81.38\scriptsize{±2.58} & \textbf{82.03\scriptsize{±1.48}}\\
        & 20  &  59.31\scriptsize{±0.27} & 56.67\scriptsize{±1.22} & 64.68\scriptsize{±0.75} & 73.54\scriptsize{±0.94} & 69.82\scriptsize{±0.71} & 73.79\scriptsize{±1.91} & 74.14\scriptsize{±1.93} & \textbf{74.81\scriptsize{±1.52}}\\
  \cmidrule(lr){1-2}\cmidrule(lr){3-7} \cmidrule(lr){8-10}

    \multirow{4}[0]{*}{{Citeseer}} & 0   & 71.81\scriptsize{±0.71} & 70.42\scriptsize{±0.39} & 74.94\scriptsize{±0.40} & 73.82\scriptsize{±0.56} & 74.92\scriptsize{±0.66} & 75.69\scriptsize{±0.69} & 78.19\scriptsize{±1.30} & \textbf{83.26\scriptsize{±0.73}} \\
        & 5   & 71.22\scriptsize{±0.61} & 68.86\scriptsize{±0.47} & 72.45\scriptsize{±0.88} & 71.41\scriptsize{±0.65} & 73.60\scriptsize{±0.45} & 75.40\scriptsize{±0.95} & 77.54\scriptsize{±1.04} & \textbf{81.66\scriptsize{±0.70}} \\
        & 10  & 67.53\scriptsize{±0.60} & 68.70\scriptsize{±0.89} & 70.16\scriptsize{±1.05} & 70.09\scriptsize{±0.48} & 73.66\scriptsize{±0.37} & 73.81\scriptsize{±1.10} & 77.04\scriptsize{±1.51} & \textbf{80.32\scriptsize{±0.49}} \\
        & 20  & 63.20\scriptsize{±1.70} & 57.95\scriptsize{±1.48} & 55.84\scriptsize{±1.28} & 67.22\scriptsize{±1.32} & 65.91\scriptsize{±1.20} & 71.06\scriptsize{±1.20}  & 72.72\scriptsize{±1.28} & \textbf{77.21\scriptsize{±0.64}}\\
  \cmidrule(lr){1-2}\cmidrule(lr){3-7} \cmidrule(lr){8-10}

    \multirow{4}[0]{*}{{Pubmed}} & 0   & 84.57\scriptsize{±0.39} & 83.25\scriptsize{±0.35} & 84.96\scriptsize{±0.08} & 84.87\scriptsize{±0.10} & {85.94\scriptsize{±0.10}} & 80.26\scriptsize{±0.43} & 85.51\scriptsize{±0.66} & \textbf{87.27\scriptsize{±0.67}}  \\
        & 5   & 81.25\scriptsize{±0.50} & 82.90\scriptsize{±0.26} & 83.00\scriptsize{±0.10} & 82.32\scriptsize{±0.11} & {83.89\scriptsize{±0.09}} & 77.86\scriptsize{±0.35} & 81.92\scriptsize{±0.59} & \textbf{83.36\scriptsize{±0.62}} \\
        & 10  & 78.96\scriptsize{±0.43} & 80.35\scriptsize{±0.21} & 80.82\scriptsize{±0.20} & 80.77\scriptsize{±0.11} & \textbf{82.13\scriptsize{±0.15}} & 76.62\scriptsize{±0.55} & 79.29\scriptsize{±0.60} & 80.99\scriptsize{±0.58} \\
        & 20  & 71.33\scriptsize{±0.40} & 73.57\scriptsize{±0.15} & 74.16\scriptsize{±0.16} & 73.41\scriptsize{±0.12} & {76.01\scriptsize{±0.19}} & 74.98\scriptsize{±0.64} & 76.87\scriptsize{±0.60} & \textbf{77.15\scriptsize{±0.56}} \\
        \cmidrule(lr){1-2}\cmidrule(lr){3-7} \cmidrule(lr){8-10}
    \multirow{4}[0]{*}{{Polblogs}} & 0   & 94.87\scriptsize{±0.19} & 95.08\scriptsize{±0.22} & {95.45\scriptsize{±0.12}} & 95.03\scriptsize{±0.57} & \textbf{95.70\scriptsize{±0.34}} & 95.42\scriptsize{±0.56} & 94.88\scriptsize{±0.43} & 94.97\scriptsize{±0.31} \\
        & 5   & 73.28\scriptsize{±0.18} & 88.86\scriptsize{±0.58} & 90.98\scriptsize{±0.69} & 90.97\scriptsize{±0.61} & 89.97\scriptsize{±1.25} & 90.18\scriptsize{±0.43} & \textbf{91.22\scriptsize{±0.38}} & 89.37\scriptsize{±0.46} \\
        & 10  & 70.91\scriptsize{±0.37} & 80.38\scriptsize{±0.85} & 85.60\scriptsize{±1.08} & 85.93\scriptsize{±1.39} & 83.66\scriptsize{±1.81} & \textbf{86.30\scriptsize{±0.70}} & 85.43\scriptsize{±0.68} & 85.07\scriptsize{±0.54} \\
        & 20  & 57.97\scriptsize{±0.41} & 55.33\scriptsize{±2.07} & 73.52\scriptsize{±0.53} & 70.47\scriptsize{±1.27} & 75.87\scriptsize{±0.88} & \textbf{82.03\scriptsize{±0.79}} & 81.96\scriptsize{±0.72} & 81.56\scriptsize{±0.18} \\
    \bottomrule
    \end{tabular}
    \caption{Classification accuracy (\%) on the perturbed graph. $p_r$ is the perturbation rate and $s_r$ is the ``seen'' rate.}
    \label{tab_accu}
\end{table*}

We assess the adversarial robustness of the 3-layer GRN under generalization demands by comparing its accuracy against other baselines on perturbed graphs. We partition a subset of the dataset as training set with proportions (also named ``seen'' rate) $s_r$ as 20\%, 40\%, and 60\%. The data within these proportions are deemed ``seen'' by the GRN, while the remaining data is categorized as ``unseen''. Utilizing Metattack as our attack approach, we adopt the standard 2-layer GCN as the surrogate model. By adjusting $p_r$, we derive the corresponding perturbed graphs. Then, we evaluate the classification performance of the baselines on these graphs, placing an emphasis on the accuracy upon model convergence. For each setting, we executed 10 iterations, tabulating both the average outcome and its variability. Table~\ref{tab_accu} reports the results.

From the data, both clean and perturbed graphs show GRN with a $s_r$ = 60\% generally surpasses the baseline in accuracy. There are three exceptions:
1) For the Pubmed dataset at $p_r$ = 10\%, this is due to EGNN using graph smoothing to enhance adversarial robustness. In this case, perturbations may be more pronounced in a certain area, and EGNN could leverage this by smoothing concentrated perturbation patterns. However, these incidents are rare, and as $p_r$ increases, GRN's accuracy returns to its peak.
2) With the Polblogs dataset at $p_r$ = 0\%, GRN is slightly behind EGNN by 0.28\%. Yet, as $p_r$ rises, the decline in GRN's accuracy is the least noticeable among all baselines, ensuring its top position.
3) An intriguing pattern emerging from the Polblogs dataset is the non-proportional relationship between the GRN's $s_r$ and its accuracy. The peculiarity of the Polblogs dataset is that its nodes lack intrinsic features. Typically, scholars have used node degrees as proxies for these absent attributes. This substitution results in the inherent attributes of Polblogs leaning towards uniformity. Expanding the training set's scale exacerbate the oversmoothing phenomenon, culminating in diminished accuracy.

\subsubsection{Ablation Studies}
GRNs combine edge-transmitted signal $\mathbf{E}^{(\ell)}_{G_i}$ and node signal $\mathbf{Z}^{(\ell)}_{G_i}$ for node $v_i$'s representation. We initiate an ablation study to delve into this process.
First, we embed only $\mathbf{Z}^{(\ell)}_{G_i}$, naming the model GRN$_{\mathbf{Z}}$. This appears similar to a 2-depth GraphSAGE with mean aggregation, indicating potential vulnerability to adversarial attacks.
We then examine the combination order of $\mathbf{E}^{(\ell)}_{G_i}$ and $\mathbf{Z}^{(\ell)}_{G_i}$. The default GRN order is GRN$_{\mathbf{E},\mathbf{Z}}$. We test GRN$_{\mathbf{Z},\mathbf{E}}$ (reversed order) and GRN$_{shuf}$ (shuffled rows).
Results (Table~\ref{tab_abl}) show GRN$_{\mathbf{Z}}$ underperforms, especially in adversarial settings, emphasizing the importance of co-embedding both signals. Precisions of GRN$_{\mathbf{Z},\mathbf{E}}$, GRN$_{\mathbf{E},\mathbf{Z}}$, and GRN$_{shuf}$ are comparable due to the edge-transmitted signal, which, combined with the node signal through shared parameters, results in consistent performance regardless of order. This suggests that GRN has the capability to recognize a latent graph structure, wherein edge-transmitted signals function as latent node signals, contributing to adversarial robustness and insensitivity to signal order.

\begin{table}[htbp]
  \centering \small

\setlength{\tabcolsep}{1.25mm}{
    \begin{tabular}{c | c | c |ccc}
    \toprule
            \multirow{2}[0]{*}{{Dataset}} & \multirow{2}[0]{*}{{$p_r$ (\%)}} & \multicolumn{1}{c |}{Standard} & \multicolumn{3}{c }{Ablated models}  \\
    \cmidrule(lr){3-3} \cmidrule(lr){4-6}

        &  & GRN$_{\mathbf{E},\mathbf{Z}}$ & GRN$_{\mathbf{Z},\mathbf{E}}$ & GRN$_{shuf}$ & GRN$_{\mathbf{Z}}$ \\
  \cmidrule(lr){1-2}\cmidrule(lr){3-6}
    \multirow{4}[0]{*}{{Cora}} & 0   &  87.75 & 87.14\scriptsize{±0.81} & 87.28\scriptsize{±0.92} & 87.74\scriptsize{±0.72} \\
        & 5   &  86.24 & 86.18\scriptsize{±0.75} & 86.34\scriptsize{±0.98} & 85.56\scriptsize{±0.87} \\
        & 10  &  82.03 & 82.81\scriptsize{±0.84} & 82.54\scriptsize{±1.05} & 79.07\scriptsize{±0.90} \\
        & 20  &  74.81 & 74.53\scriptsize{±0.80} & 74.42\scriptsize{±1.36} & 63.54\scriptsize{±1.30} \\
  \cmidrule(lr){1-2}\cmidrule(lr){3-6}

    \multirow{4}[0]{*}{{Citeseer}} & 0   & 83.26 & 83.07\scriptsize{±0.64} & 83.52\scriptsize{±0.92} & 82.23\scriptsize{±0.97} \\
        & 5   & 81.66 & 81.49\scriptsize{±0.71} & 81.45\scriptsize{±0.95} & 79.47\scriptsize{±1.10} \\
        & 10  & 80.32 & 80.10\scriptsize{±0.65} & 80.46\scriptsize{±1.17} & 74.39\scriptsize{±1.44} \\
        & 20  & 77.21 & 76.95\scriptsize{±1.26} & 76.84\scriptsize{±1.53} & 69.05\scriptsize{±1.86} \\

    \bottomrule
    \end{tabular}}
    \caption{Classification accuracy (\%) of ablated models.}
    \label{tab_abl}
\end{table}

\section{Conclusions}
We addressed critical concerns surrounding the transductive nature of existing robust graph learning models. We began by establishing the equivalence between GCN-driven MP and edge-based waves. Subsequently, we demonstrated that a 3-layer GCN capitalizes on the unique resonance intrinsic to waves to achieve robustness. Delving deeper, we formalized this resonance as a coupling between a node and its surrounding local structure. We then introduced an inductive graph learning model, GRN. Experimental results not only corroborated our theoretical insights but also highlighted the exemplary robustness of the proposed GRN model.

\section{Acknowledgments}
This work is supported in part by the National Key Research and Development Program of China (No. 2020YFB1805400), the National Natural Science Foundation of China (No. U19A2068, No. 62372313), and the Sichuan Science and Technology Program (No. 2023YFG0113).

\begin{quote}
\bibliography{aaai24}
\end{quote}

\clearpage
\appendix

\subsection{Proofs}

\subsubsection{Proof of Theorem~\ref{thm_1}}

Node signals vibrate from high to low dimensions during MP, which is implemented through embedding. To ensure the uniformity of node signal dimensions during the MP process, we introduce a invertible transfer function, denoted as $\mathcal{T}_{d_0}:\mathbb{R}^{d_k} \to \mathbb{R}^{d_0}, \forall d_k \leq d_0$, to standardize the dimensions of $\mathbf{Z}^{(\ell)}$ and $\mathbf{Z}^{(0)}$. This function can extend the process of the $k$-th MP to $\mathbf{Z}^{(k-1)} \stackrel{\text{MP}}{\longrightarrow} \mathbf{Z}^{(k)}  \stackrel{\mathcal{T}_{d_0}}{\longrightarrow} {\bar{\mathbf{Z}}^{(k)}}$, where ${\bar{\mathbf{Z}}^{(k)}}=[\mathbf{Z}^{(k)}|\mathbf{0}_{N\times (d_0-d_k)}]$. This process can be interpreted as all embeddings induced by MP are completed in $\mathbb{R}^{d_0}$. i.e., embedding into $d_k$ dimensions at the k-th MP, and the subsequent $d_0-d_k$ dimensions collapse to zero. Indeed, with this approach, MP can be viewed as signal propagation in a space of equal dimensions.

The edge-based Laplacian wave equation is (also introduced in main text):
\begin{equation}\label{eq_graph_wave}
    \frac{\partial^2 g}{\partial \ell^2}=-\mathbf{L}^{\ell}g \cdot c,
\end{equation}
 where $c$ is a constant. In this context, we instantiate $g$ as the signal ${\bar{\mathbf{Z}}^{(\ell)}}$.
Then we define $c$ as a signal amplitude matrix:
\begin{myDef}
The amplitude in $k$-th MP is $\mathbf{S}_k \in \mathbb{R}^{d_0 \times d_0}$ represents the transmission amplitude of signals in different dimensions. It is a variable matrix that defines the adjustable parameters of the graph-based function at different times. It performs an equal-dimensional transformation of the signal through right multiplication.
\end{myDef}
The initial speed of the first-hop MP can thus be denoted as $\mathbf{S}_0$.
Consequently, we derive that $\frac{\partial^2 \bar{\mathbf{Z}}^{(\ell)}}{\partial \ell^2} =  \mathbf{L}^{\ell}\bar{\mathbf{Z}}^{(0)} \mathbf{S}_{\ell} $.
Hence, the element-wise fluctuation speed~[1] of the signal $\bar{\mathbf{Z}}^{(\ell)}$ , denoted as $\mathbf{V}_k\in\mathbb{R}^{N\times d_0}$, can be described, and it has the relationship
\begin{equation}\label{eq_speed}
\mathbf{V}_{\ell} = \frac{\partial \bar{\mathbf{Z}}^{(\ell)}}{\partial \ell} = \mathbf{L}^{\ell}\bar{\mathbf{Z}}^{(0)} \int  \mathbf{S}_{\ell} d\ell,
\frac{\partial \mathbf{V}_{\ell}  }{\partial \ell}=\mathbf{L}^{\ell}\bar{\mathbf{Z}}^{(0)} \mathbf{S}_{\ell}.
\end{equation}
We regard $\bar{\mathbf{Z}}^{(\ell)}$ as a sampling within the continuous signal with a minimal step $h$, the wave described by $\bar{\mathbf{Z}}^{(\ell)}$ can thus be formalized as
\begin{equation}\label{eq_wave_signal}
\bar{\mathbf{Z}}^{(k+h)} = \bar{\mathbf{Z}}^{(k)} + h\mathbf{V}_k,
\mathbf{V}_{k+h}=\mathbf{V}_k+h\mathbf{L}^{k}\bar{\mathbf{Z}}^{(0)} \mathbf{S}_{k}.
\end{equation}
One may write
\begin{align}\label{eq_wave_chain}
  \bar{\mathbf{Z}}^{(k+h)} &= \bar{\mathbf{Z}}^{(k)} + \mathbf{V}_{k-h} + \mathbf{L}^{k}\bar{\mathbf{Z}}^{(0)} \mathbf{S}_{k-h}   \notag \\
  &=\bar{\mathbf{Z}}^{(k)} + \mathbf{V}_{k-2h} + \mathbf{L}^{k}\bar{\mathbf{Z}}^{(0)} \mathbf{S}_{k-2h} + \mathbf{L}^{k}\bar{\mathbf{Z}}^{(0)} \mathbf{S}_{k-h},
\end{align}
and this discrete representation iterating with a step size of $h$ can be written in a continuous expression:
{\small
\begin{align}\label{eq_wave_chain2}
 \lim\limits_{h \to 0}\bar{\mathbf{Z}}^{(k+h)} &= \bar{\mathbf{Z}}^{(k)}  + \mathbf{V}_0 + \mathbf{L}^{k}\bar{\mathbf{Z}}^{(0)}\int_{1}^{k-1}\mathbf{S}_{\ell} d\ell \notag \\
  &=\bar{\mathbf{Z}}^{(k)}  + \mathbf{L}^{k}\bar{\mathbf{Z}}^{(0)}  \left(\int_0^1  \mathbf{S}_{\ell} d\ell + \int_{1}^{k-1}\mathbf{S}_{\ell} d\ell \right)\notag \\
  &=\bar{\mathbf{Z}}^{(k)}  + \mathbf{L}^{k}\bar{\mathbf{Z}}^{(0)}   \int_0^{k-1}  \mathbf{S}_{\ell} d\ell.
\end{align}
}
In order to establish a direct correlation between $\bar{\mathbf{Z}}^{(k+h)}$ and $\bar{\mathbf{Z}}^{(0)}$, instead of via a chain-like propagation through $\bar{\mathbf{Z}}^{(k)},\bar{\mathbf{Z}}^{(k-h)},\ldots$, we introduce the Moore-Penrose (MP) pseudoinverse~[2] to simplify Equation~\eqref{eq_wave_chain}. Specifically, we introduce a dimension matrix $\mathbf{C}_k \in \mathbb{R}^{d_0 \times d_0}$, which is associated with $\mathbf{S}_k$, calculated by
\begin{equation}\label{eq_Ck}
  \mathbf{C}_k  = \left( \mathbf{L}^{k}\bar{\mathbf{Z}}^{(0)} \right)^{\dagger}\left(\mathbf{L}^{k}\bar{\mathbf{Z}}^{(k)} + \mathbf{L}^{k}\bar{\mathbf{Z}}^{(0)} \int_0^{k-1}  \mathbf{S}_{\ell} d\ell \right).
\end{equation}
Then, a direct connection between $\bar{\mathbf{Z}}^{(k+1)}$ and $\bar{\mathbf{Z}}^{(0)}$ can be obtained as\
\begin{equation}\label{eq_Zk_Z0}
  \bar{\mathbf{Z}}^{(k)} = \mathbf{L}^{k}\bar{\mathbf{Z}}^{(0)} \mathbf{C}_k,
\end{equation}
which delineates the general term formula for equal-dimensional waves $\bar{\mathbf{Z}}^{(\ell)}$. Given that $\mathbf{Z}^{\ell}=\mathcal{T}_{d_0}^{-1}(\bar{\mathbf{Z}}^{(\ell)})$, the proposition henceforth is such that, if one were to establish the validity of $\mathcal{T}_{d_0}^{-1}(\mathbf{L}^{k}\bar{\mathbf{Z}}^{(0)}\mathbf{C}_k)$, signifying that a $\mathbf{C}_k$ exists that would embedding the trailing $d_0-d_k$ columns of $\mathbf{L}^{k}\bar{\mathbf{Z}}^{(0)}$ to be minimal value matrices (denoted as $\mathbf{T}_{d_k|0}=[\mathbf{T}_{d_k}| \mathbf{0}_{N \times (d_0-d_k)}]$), then a inverse mapping could be constructed that would allow for the transition from equal-dimensional waves $\bar{\mathbf{Z}}^{(\ell)}$ to unequal-dimensional waves $\mathbf{Z}^{(\ell)}$.
To address this problem, we split $\mathbf{C}_k$ into two segments: a left $d \times (d_k)$ matrix $\mathbf{C}_{k,L}$ and a right $d \times (d_0-d_k)$ matrix $\mathbf{C}_{k,R}$.
Given that, we have: $\mathbf{L}^{k}\bar{\mathbf{Z}}^{(0)}\mathbf{C}_{k} = [\mathbf{L}^{k}\bar{\mathbf{Z}}^{(0)}\mathbf{C}_{k,L}|\mathbf{L}^{k}\bar{\mathbf{Z}}^{(0)}\mathbf{C}_{k,R}]$.
Since $\bar{\mathbf{Z}}^{(0)}={\mathbf{Z}}^{(0)}$, it thus holds that $\bar{\mathbf{Z}}^{(0)}$ is of full column rank.
Consequently, we may conclude that for any arbitrary real matrix $\mathbf{T}_{d_k}$, there exist $\mathbf{C}_{k,L}$ such that $\mathbf{L}\bar{\mathbf{Z}}^{(0)}\mathbf{C}_{k,L}=\mathbf{T}_{d_k}$, while supporting
$\mathbf{L}^{k}\bar{\mathbf{Z}}^{(0)}\mathbf{0}_{N\times (d_0-d_k)}=\mathbf{0}_{N\times (d_0-d_k)}$ holds.
 Therefore, there exists a matrix
 \begin{equation}\label{eq_C}
\mathbf{C}_{k}=[\mathbf{C}_{k,L}|\mathbf{0}_{N\times (d_0-d_k)}],
 \end{equation}
such that $\mathbf{L}\bar{\mathbf{Z}}^{(0)}\mathbf{C}_{k}=[\mathbf{T}_{d_k}|\mathbf{0}_{N\times (d_0-d_k)}]=\mathbf{T}_{d_k|0}$, thereby confirming the validity of $\mathcal{T}_{d_0}^{-1}$.
Hence, Equation~\eqref{eq_Zk_Z0} can be expressed inversely as
\begin{equation}\label{eq_wave_inverse}\small
  {\mathbf{Z}}^{(k+1)} = \mathcal{T}_{d_0}^{-1}(\bar{\mathbf{Z}}^{(k+1)}) = \mathbf{L}^{k}{\mathbf{Z}}^{(0)}\mathcal{T}_{d_0}^{-1}({\mathbf{C}_{k}}) = \mathbf{L}^{k}{\mathbf{Z}}^{(0)}\mathbf{C}_{k,L}.
\end{equation}
Taking into account the dimension of variation under the wave, and the independent amplitude correlation when the signal fluctuates once, we apply rank-$\eta$ truncated singular value decomposition ($\eta$-TSVD)~[3] to $\mathbf{C}_{k,L}$ to approximately obtain $k$ amplitude-correlated matrices. Given $\forall d_0 \leq \eta \leq d_k $, $\eta$-TSVD decomposes $\mathbf{C}_{k,L}$ as
$\mathrm{SVD_T^{(\eta)}} (\mathbf{C}_{k,L})=\mathbf{U}_k \mathbf{\Lambda}_k \mathbf{V}_k$,
where $\mathbf{U}_k\in \mathbb{R}^{d_0 \times \eta}$, $\mathbf{\Lambda}_k \in \mathbb{R}^{\eta \times \eta}$ , and $\mathbf{V}_k \in \mathbb{R}^{\eta \times d_k}$ are matrices whose columns are the first $\eta$ left singular vectors, the first $\eta$ singular values, and the first $\eta$ right singular vectors of the $\mathbf{C}_{k,L}$, respectively. We denote:
1) $\mathbf{U}_k \mathbf{\Lambda}_k = \widehat{\mathbf{C}}_{k-1} \in \mathbb{R}^{d_0 \times \eta}$;
2) $\widehat{\mathbf{U}}_{k-1}$, $\widehat{\mathbf{\Lambda}}_{k-1}$ and $\widehat{\mathbf{V}}_{k-1}$ as the same of the above definitions;
3) $\widehat{\mathbf{C}}_{k-2} = \widehat{\mathbf{U}}_{k-1} \widehat{\mathbf{\Lambda}}_{k-1}$.
Since TSVD can decompose any real-valued matrix for any valid $\eta$, we apply continuous $\eta$-TSVD to $\mathbf{C}_{k,L}$, yielding:
\begin{align}\label{eq_TSVD_chain}
  \mathrm{SVD_T^{(\eta_k)}} (\mathbf{C}_{k,L}) &= \widehat{\mathbf{C}}_{k-1}  \mathbf{V}_k \notag \\
  \mathrm{SVD_T^{(\eta_{k-1})}} (\widehat{\mathbf{C}}_{k-1}) &= \widehat{\mathbf{C}}_{k-2 } \mathbf{V}_{k-1} \notag \\
  \ldots \notag \\
  \mathrm{SVD_T^{(\eta_{2})}} (\widehat{\mathbf{C}}_{2}) &= \widehat{\mathbf{C}}_{1 } \mathbf{V}_{2}
\end{align}
Since TSVD provides an approximate equality with introducing a minimum error term:
\begin{equation}\label{eq_OGamma}\small
O(\Gamma) = O\left( \sum_{i=2}^{k-1} || \widehat{\mathbf{C}}_{i} - \widehat{\mathbf{C}}_{i-1} \mathbf{V}_{i} ||_F + ||\mathbf{C}_{k,L} - \widehat{\mathbf{C}}_{k-1}  \mathbf{V}_k ||_F  \right),
\end{equation}
we can summarize Equation~\eqref{eq_TSVD_chain}, decomposing $\mathbf{C}_{k,L}$ into a form of $k$ continuous multiplication, to formalize the $k$ amplitude matrices introduced by $k$ iterations of MP:
\begin{equation}\label{eq_SVD_Final}
  \mathbf{C}_{k,L} = \widehat{\mathbf{C}}_{1} \prod_{i=2}^{k} \mathbf{V}_{k} + O(\Gamma).
\end{equation}
Substituting Equation~\eqref{eq_SVD_Final} into Equation~\eqref{eq_wave_inverse}, we obtain:
\begin{equation}\label{eq_Zk_Final}
  \mathbf{Z}^{(k+1)} = \mathbf{L}^{k}{\mathbf{Z}}^{(0)} \widehat{\mathbf{C}}_{1} \prod_{i=2}^{k} \mathbf{V}_{k}
\end{equation}
Now, to endow the fluctuation equation represented by Equation~\eqref{eq_Zk_Final} with the capability of backpropagation, we introduce $k$ nonlinear activation function $\sigma_1,\ldots \sigma_k$,  Subsequently, we instantiate $\widehat{\mathbf{C}}_{1}, \mathbf{V}_{2}, \mathbf{V}_{3},\ldots,\mathbf{V}_{k}$ as $k$ trainable matrices. As we expounded in the preceding text, the dimensions of these matrices merely need to satisfy the condition of input-output decrement, and thus these dimensions can be arbitrarily set. We denote X as the $i$-th trainable matrix by given a decrement dimension $d_0, \eta_k, \eta_{k-1},\ldots, \eta_3, \eta_{2}, d_k$, the aforementioned matrices can be instantiated as $\mathbf{W}_1=\widehat{\mathbf{C}}_{1}\in\mathbb{R}^{d_0 \times \eta_k}, \mathbf{W}_2=\mathbf{V}_{k}\in\mathbb{R}^{ \eta_k \times \eta_{k-1}},\ldots \mathbf{W}_k = \mathbf{V}_{2}\in\mathbb{R}^{ \eta_2 \times d_k}$.
In summary, by making Equation~\eqref{eq_Zk_Final} trainable, we obtain the following form of a trainable wave equation:
\begin{multline}\label{eq_trainable_wave}
\mathbf{Z}^{(k)} = \mathrm{W{\scriptstyle AVE}}(\mathbf{Z},k;\mathbf{L}) \\
= \sigma_k \underbrace{ \Big( \mathbf{L} \ldots \overbrace{\sigma_2 (\mathbf{L}
\underbrace{\sigma_1 ( \mathbf{L}{\mathbf{Z}}^{(0)}\mathbf{W}_1 )}_{\substack {\text{layer }1 \\ \ldots}} \mathbf{W}_2 )}^{\text{layer }2}\ldots \mathbf{W}_k \Big) } _{\text{layer }k}
\\= \mathcal{M}(\mathbf{Z},k;\mathbf{L}).
\end{multline}
We note that Equation~\eqref{eq_trainable_wave} is the trainable form of the edge-based Laplacian wave equation (Equation~\eqref{eq_graph_wave}), and it is also the forward propagation form of the GCN. Through the derivation process we have provided, the equivalence between the GCN and the wave equation is thus substantiated.

\subsubsection{Proof of Theorem~\ref{thm_2}}

We first gives the vectorized representation of the resonance intensity. For each pair of nodes $(v_j,v_k)$ in $\mathcal{N}_i$, if there is an edge between $v_j$ and $v_k$, then this edge is counted in $T_i$. This is indicated by $\mathbf{A}_{j,k}=1$. Therefore $T_i$ equals the sum of $\mathbf{A}_{j,k}$ over all nodes pairs $(v_j,v_k)$ in $\mathcal{N}_i$. Then, $T_i$ can be obtained by three steps:
\begin{itemize}
  \item First, the number of edges on all nodes pairs of $\mathcal{G}$ $(v_j,v_k)$ (not just pairs of neighbors of node $v_i$) is $\sum_{v_j, v_k \in \mathcal{N}_i} \mathbf{A}_{j,k}$.
  \item Then, to ensure that $v_j , v_k \in \mathcal{N}_i$, we need to include the terms $A_{i,j}$ and $\mathbf{A}_{k,i}$ in the above summation. This is because $\mathbf{A}_{i,j} = \mathbf{A}_{k,i} =1$ if and only if $v_j$ and $v_k$ are both neighbors of node $v_i$, i.e., $\mathbf{A}_{i,j}\mathbf{A}_{j,k}\mathbf{A}_{k,1}=1$ if and only if $v_i$, $v_j$ and $v_k$ form a triangle.
  \item Finally, $\sum_{i,j}\mathbf{A}_{i,j}\mathbf{A}_{j,k}\mathbf{A}_{k,1}$ counts the number of triangles that include node $v_i$, This gives that $T_i=\sum_{i,j} \mathbf{A}_{i,j}\mathbf{A}_{j,k}\mathbf{A}_{k,i}$
\end{itemize}
Then, we have

{\small
\begin{multline}\label{eq_LCC_chain}
 \bar{{z}}_{i}^{(0)} T_i =  \bar{{z}}_{i}^{(0)} \sum_{v_j, v_k \in \mathcal{N}_i} \mathbf{A}_{jk}=  \sum_{j,k} \mathbf{A}_{i,j}\mathbf{A}_{j,k}\mathbf{A}_{k,i} \sum_{l} \mathbf{Z}_{i,l}  \\
=  \sum_{j} \mathbf{A}_{i,j}  \left( \mathbf{A}\mathbf{A}^{\top} \right)_{j,i} \sum_{l} \mathbf{Z}_{i,l}.
\end{multline}
}
Hence, according to Definition~\ref{def_RI}, one may write
{\small
\begin{multline}\label{eq_LCC_chain2}
R(v_i;0)  = \bar{{z}}_{i}^{(0)} T_i + 2  p_i + 8 \mathrm{deg}_i  \\
= \sum_{j} \mathbf{A}_{i,j}  \left( \mathbf{A}^2 \right)_{j,i} \sum_{l} \mathbf{Z}_{i,l} + 2\sum_{j}(\mathbf{A}^2)_{i,j} + 8 \sum_{j}\mathbf{A}_{i,j} .
\end{multline}
}
We denote by $[1]_{N \times 1}\in\mathbb{R}^{N  \times 1}$ whose entries are all equal to $1$.
We utilize right multiplication by $[1]_{1 \times N}$ for row-wise summation of the matrix.
Subsequently, we engage in the augmentation of $R(\cdot)$ to its vectorized representation:
\begin{equation}\label{eq_LCC_chain3}\small
\mathbf{R}'(v_i;0) = ( \mathbf{A}^3 \mathbf{Z_0}[1]_{N \times 1} + 2 \mathbf{A}^2[1]_{N \times 1} + 8\mathbf{A}[1]_{N \times 1})_i
\end{equation}
Here, $\mathbf{Z_0}=[\mathbf{Z}|\mathbf{0}] \in \mathbb{R}^{N \times N}$ represents the expansion of $\mathbf{Z}$ into a square matrix $\mathbf{Z_0}$ by padding zeros on the right side of $\mathbf{Z}$. Moving forward, we incorporate the Sigmoid function $S(\cdot)$ and elucidate its role through the matrix notation
\begin{equation}\label{eq_matrix_notat}
  \mathbf{B} = S(\mathbf{A}S(\mathbf{A}\mathbf{Z})), \mathbf{C} = S(\mathbf{A}\mathbf{Z}).
\end{equation}
Then we stipulate the minimum error function as
\begin{equation}\label{eq_O}
E(\mathbf{X})= -\frac{1}{288}\mathbf{X}^3 + O(\mathbf{X}^5),
\end{equation}
and denote following minimum error terms $\mathbf{o}_1 = E(\mathbf{A}\mathbf{Z})$, $\mathbf{o}_2 = E(\mathbf{A}\mathbf{C})$, $\mathbf{o}_3 = E(\mathbf{A}\mathbf{B})$ and $\mathbf{o} = \frac{1}{16}\mathbf{A}^2\mathbf{o}_1[1]_{N\times 1} +\frac{1}{4}\mathbf{A} \mathbf{o}_2 [1]_{N\times 1} + \mathbf{o}_3[1]_{N\times 1}$.
Utilizing the aforementioned notation, we can deduce
{\small
\begin{align}\label{eq_Build1}
& \frac{1}{64}\left( \mathbf{A}^3 \mathbf{Z_0}[1]_{N \times 1} +  2\mathbf{A}^2[1]_{N \times 1} + 8\mathbf{A}[1]_{N\times 1} \right) + \mathbf{o} + [\frac{1}{2}]_{N \times 1} \notag \\
&= [\frac{1}{2}]_{N \times 1} + \frac{1}{64}\mathbf{A}^3 \mathbf{Z_0} [1]_{N \times 1} + \frac{1}{16} \mathbf{A}^2[\frac{1}{2}]_{N \times 1} + \frac{1}{4}\mathbf{A}[\frac{1}{2}]_{N \times 1} + \mathbf{o}  \notag \\
&= [\frac{1}{2}]_{N \times 1}  +  \frac{1}{4} (\frac{1}{4})^2\mathbf{A}^3 \mathbf{Z_0}[1]_{N \times 1} + \frac{1}{4} \frac{1}{4} \mathbf{A}^2[\frac{1}{2}]_{N \times 1} + \frac{1}{4}\mathbf{A}[\frac{1}{2}]_{N \times 1} \notag \\
&\quad +  \frac{1}{4} \frac{1}{4} \mathbf{A}^2\mathbf{o}_1[1]_{N \times 1} + \frac{1}{4}\mathbf{A}\mathbf{o}_2[1]_{N \times 1} +\mathbf{o}_3[1]_{N\times 1}   \notag \\
&= [\frac{1}{2}]_{N \times 1} + \frac{1}{4}\mathbf{A} \Big( [\frac{1}{2}]_{N \times 1} + \frac{1}{4}\mathbf{A}[\frac{1}{2}]_{N \times 1}  + (\frac{1}{4})^2 \mathbf{A}^2  \mathbf{Z_0}[1]_{N \times 1} \notag \\
&\quad+ \frac{1}{4}\mathbf{A} \mathbf{o}_1[1]_{N \times 1} + \mathbf{o}_2[1]_{N \times 1} \Big) + \mathbf{o}_3[1]_{N \times 1} \notag \\
&= [\frac{1}{2}]_{N \times 1} + \frac{1}{4}\mathbf{A} \bigg( [\frac{1}{2}]_{N \times 1} + \frac{1}{4} \mathbf{A} \Big( [\frac{1}{2}]_{N\times 1} + \frac{1}{4} \mathbf{A} \mathbf{Z_0} [1]_{N \times 1} \notag \\
&\quad + \mathbf{o}_1[1]_{N\times 1} \Big) +\mathbf{o}_2[1]_{N\times 1} \bigg) +\mathbf{o}_3[1]_{N\times 1}.
\end{align}
}
For Equation~\eqref{eq_Build1}, we extract $[1]_{N \times 1}$ (there are countless methods of extraction, this is because the right multiplication by $[1]_{N \times 1}$ gives $[\frac{1}{2}]_{N \times 1}$, indicating that the sum of each row of an $N\times N$ matrix is $\frac{1}{2}$. Here, we consider the case where all elements of this matrix have the average value $\frac{1}{2N}$), thereby deriving Equation~\eqref{eq_Build1} into:
{\small
\begin{align}\label{eq_Build1_5}
&\frac{1}{64}\left( \mathbf{A}^3 \mathbf{Z_0}[1]_{N \times 1} +  2\mathbf{A}^2[1]_{N \times 1} + 8\mathbf{A}[1]_{N\times 1} \right) + \mathbf{o} + [\frac{1}{2}]_{N \times 1} \notag \\
&= \bigg([\frac{1}{2N}]_{N \times N} + \frac{1}{4}\mathbf{A} \bigg( [\frac{1}{2N}]_{N \times N} + \frac{1}{4} \mathbf{A} \Big( [\frac{1}{2N}]_{N\times N} + \frac{1}{4} \mathbf{A} \mathbf{Z_0} \notag \\
&\quad + \mathbf{o}_1 \Big) +\mathbf{o}_2 \bigg) +\mathbf{o}_3\bigg) [1]_{N \times 1}.
\end{align}
}

The matrix-based Taylor series expansion~[4] for $\mathbf{X}$ is
\begin{multline}\label{eq_Tylor_1}
S(\mathbf{X}) = [\frac{S(0)}{N}]_{N\times N} + S'(0)\mathbf{X} + \frac{1}{2} \mathbf{X}^2 S''(0) + \\ \frac{1}{6} \mathbf{X}^3 S'''(0)  + \frac{1}{24} \mathbf{X}^4 S''''(0) + O(\mathbf{X})^5).
\end{multline}
Since $S(0)=\frac{1}{2}$, $S'(0)=\frac{1}{4}$, $S''(0)=0$, $S'''(0)=-\frac{1}{48}$ and $S''''(0)=0$, by substituting them into Equation~\eqref{eq_Tylor_1}, we have
{\small
\begin{align}\label{eq_Tylor_2}
 S(\mathbf{X}) &= [\frac{1}{2N}]_{N\times N} + \frac{1}{4}\mathbf{X} + 0 \frac{1}{2}\mathbf{X}^2  - \frac{1}{48} \frac{1}{6} \mathbf{X}^3 + 0 \frac{1}{24} \mathbf{X}^4 + O(\mathbf{X}^5) \notag \\
 &= [\frac{1}{2N}]_{N\times N} + \frac{1}{4}\mathbf{X} - \frac{1}{288} \mathbf{X}^3 + O(\mathbf{X}^5) \notag \\
 &= [\frac{1}{2N}]_{N\times N} + \frac{1}{4}\mathbf{X} + E(\mathbf{X}).
\end{align}
}

Hence, Equation~\eqref{eq_Build1}, through astute algebraic manipulation, can be metamorphosed into the form of a Taylor series expansion, as delineated by Equation~\eqref{eq_Tylor_2}. In the ensuing steps, we shall meticulously engage in the transformation of the discrete constituents within Equation~\eqref{eq_Build1}, with the aim of perpetuating the advancement of its derivation
{ \small
\begin{align}\label{eq_Tylor_items}
&[\frac{1}{2N}]_{N\times N} + \frac{1}{4} \mathbf{A}  + \mathbf{o}_1 \notag \\
&\qquad= [\frac{1}{2N}]_{N\times N} + \frac{1}{4} \mathbf{A} \mathbf{Z_0} + E(\mathbf{A}\mathbf{Z_0})=S(\mathbf{A}\mathbf{Z_0})=\mathbf{C}, \notag \\
&[\frac{1}{2N}]_{N\times N} + \frac{1}{4} \mathbf{A} \Big( [\frac{1}{2N}]_{N\times N}+ \frac{1}{4} \mathbf{A} \mathbf{Z_0} + \mathbf{o}_1 \Big) + \mathbf{o}_2 \notag \\
&\qquad= [\frac{1}{2N}]_{N\times N} + \frac{1}{4} \mathbf{A} \mathbf{C} + \mathbf{o}_2 \notag \\
&\qquad= [\frac{1}{2N}]_{N\times N} + \frac{1}{4} \mathbf{A} \mathbf{C}  + E(\mathbf{A}\mathbf{C}) = S(\mathbf{A}\mathbf{C}) = \mathbf{B}, \notag \\
& [\frac{1}{2N}]_{N \times N} + \frac{1}{4}\mathbf{A} \bigg( [\frac{1}{2N}]_{N \times N} + \frac{1}{4} \mathbf{A} \Big( [\frac{1}{2N}]_{N\times N} + \frac{1}{4} \mathbf{A} \mathbf{Z_0} \notag \\
& + \mathbf{o}_1 \Big) +\mathbf{o}_2 \bigg) +\mathbf{o}_3 = [\frac{1}{2N}]_{N\times N} + \frac{1}{4}\mathbf{A} \mathbf{B} + \mathbf{o}_3 \notag \\
&\qquad=  S(\mathbf{A}\mathbf{B}) = S(\mathbf{A}S(\mathbf{A}S(\mathbf{A}\mathbf{Z_0}))).
\end{align}
}

Substituting Equation~\eqref{eq_Tylor_items} into Equation~\eqref{eq_Build1_5}, we can ultimately derive Equation~\eqref{eq_Build1} into
\begin{align}\label{eq_Tylor_final_0}
  &\frac{1}{64}\left( \mathbf{A}^3 \mathbf{Z_0}[1]_{N \times 1} +  2\mathbf{A}^2[1]_{N \times 1} + 8\mathbf{A}[1]_{N\times 1} \right) + \mathbf{o} + [\frac{1}{2}]_{N \times 1} \notag  \\
    & \qquad = S(\mathbf{A}S(\mathbf{A}S(\mathbf{A}\mathbf{Z_0}))) [1]_{N \times 1} \notag \\
    & \qquad = \mathrm{\scriptstyle SUM_{row}}(S(\mathbf{A}S(\mathbf{A}S(\mathbf{A}\mathbf{Z_0})))_{i,:}),
\end{align}
where $\mathrm{\scriptstyle SUM_{row}}(\cdot)$ denotes the operation of summing the elements of each row in the input matrix.
For $\mathbf{Z_0}$, irrespective of left multiplication by any matrix or application of an element-wise activation function, its rightmost $N-d_0$ columns persist as zero. As a result, when right-multiplied by $[1]_{N \times 1}$, (which effectively sums each row), the outcomes for $\mathbf{Z_0}$ and $\mathbf{Z}$ are identical. This is because the zeroed columns in $\mathbf{Z_0}$ do not contribute to the row sum. Consequently, even when $\mathbf{Z_0}$ is reverted back to $\mathbf{Z}$, Equation~\eqref{eq_Tylor_final_0} still holds. Then, it is perceptible that the yield of Equation~\eqref{eq_Tylor_final_0} is emblematic of a 3rd MP scheme, propelled by the GCN — wherein the GCN variant deployed abstains from the consideration of self-loops, thereby adopting the use of matrix $\mathbf{A}$ in lieu of $\mathbf{L}$ — and integrates the sigmoid function as its activation. Therefore, we can write $\mathcal{M}(\mathbf{Z},3;\mathbf{A}) = S(\mathbf{A}S(\mathbf{A}S(\mathbf{A}\mathbf{Z})))$.
Consequently, by effectuating the incorporation of Equation~\eqref{eq_Tylor_items} into Equations~\eqref{eq_Build1} and~\eqref{eq_LCC_chain3}, and electing to neglect the error magnitudes $\mathbf{o}$, we are enabled to derive the following expression
\begin{align}\label{eq_R_M_relation}
  \mathbf{R}'(v_i;0) &\propto 64 \mathrm{\scriptstyle SUM_{row}} (\mathcal{M}(\mathbf{Z},3;\mathbf{A})_{i,:}) - [32]_{N\times 1}.
\end{align}
Similarly, we have
\begin{align}\label{eq_LCC_chain4}
\mathbf{R}'(v_i;1) &\propto 64 \mathrm{\scriptstyle SUM_{row}} (\mathcal{M}(\mathbf{Z}^{(1)},3;\mathbf{A}))_{i,:} - [32]_{N\times 1} \notag \\
&= 64 \mathrm{\scriptstyle SUM_{row}} \mathcal{M}(\mathbf{Z},4;\mathbf{A}))_{i,:} - [32]_{N\times 1}.
\end{align}
Hence, we can write
\begin{align}\label{eq_R_M_general}
  \mathbf{R}'(v_i;k) &\propto 64 \mathrm{\scriptstyle SUM_{row}} (\mathcal{M}(\mathbf{Z},k+3;\mathbf{A}))_{i,:} - [32]_{N\times 1} \notag \\
  &= 64[\bar{{z}}_{i}^{(k+3)}]_{N\times 1}-[32]_{N\times 1}.
\end{align}
Through the transposition of Equation~\eqref{eq_R_M_general} into a scalar representation (i.e., $R(v_i;k)$), we are capacitated to consummate the derivation of Equation~\eqref{eq_thm_2}.

\subsubsection{Proof of Theorem~\ref{thm_cost}}

Adversaries employ the following methodology to attack node $v_i$: Initially, a target label $\mathbf{y}'_i$ is defined, wherein $\mathbf{y}'_i$ and $\mathbf{y}_i$ differ only in the label value of the target nodes (regardless of single target or multiple targets), which is altered to the target labels, while the labels of all other nodes remain unchanged. Subsequently, by modifying the graph structure, specifically $\mathbf{A}$, with a budget constraint $r$ a new graph which with $\mathbf{A}'$ as adjacency matrix is obtained. The objective is to maximize the similarity between the output and $\mathbf{y}'_i$ (i.e., minimize the loss between them) after the forward propagation by $\mathcal{M}$.

Ultimately, as the primary goal is to manipulate the one-hot output~[5], i.e., the magnitude relationship in the $i$-th row of $\mathbf{Z}^{(K)}$, an activation function is added only in the final layer to compute the loss function with the label matrix~[6]. Consequently, the optimization target of the adversaries can be expressed as:
\begin{equation}\label{eq_adver_goal_ini}
  \arg\min_{\mathbf{A}'} \mathrm{MSE}(\sigma(\mathbf{A}'^K\mathbf{Z}\prod_{i=1}^{K}\mathbf{W}^{(i)}),\mathbf{Y}').
\end{equation}
For $r$, less than 5\% of total edges is considered to be inconspicuous in common. Hence, given the necessity to minimize the perceptibility of the perturbation, $r<5\%|\mathcal{V}|$. Therefore, by introducing a sparse matrix $\mathbf{P_1}$ containing only 0, 1, and -1, the perturbation process can be represented as $\mathbf{A}' = \mathbf{A} + \mathbf{P_1}$.  Simultaneously, during training, as the activation functions between layers have been disregarded (as aforementioned, this does not affect the size relationship of the output), the parameter matrices between different layers can be unified into a single parameter matrix. We denote $\mathbf{W}_K=\prod_{i=1}^{K}\mathbf{W}^{(i)}$, hence, Equation~\eqref{eq_adver_goal_ini} can be rewritten as:
\begin{equation}\label{eq_adver_goal_simply}
  \arg\min_{\mathbf{P_1}} \mathrm{MSE}(\sigma((\mathbf{A}+\mathbf{P_1})^K\mathbf{Z}\mathbf{W}_K),\mathbf{Y}').
\end{equation}
Considering when $K=1$, Equation~\eqref{eq_adver_goal_simply} becomes:
\begin{equation}\label{eq_adver_goal_K_1}
  \arg\min_{\mathbf{P_1}} \mathrm{MSE}(\sigma(\mathbf{A}\mathbf{Z}\mathbf{W}^{(1)} + \mathbf{P_1}\mathbf{Z}\mathbf{W}^{(1)}),\mathbf{y'}).
\end{equation}

To quantify the cost of the equation above, we consider that in non-adversarial scenarios, the GCN can reach the computational universality of the ground-truth labels, i.e., there always exists $\mathbf{W}$ and a minimal value $\sigma$ such that $\mathrm{MSE}(\sigma(\mathbf{A}\mathbf{Z}\mathbf{W}),\mathbf{Y}) \leq \epsilon$. Since $\mathbf{P_1}$ is perturbable, to maintain the concealment of the perturbations and minimize the modification degree of the GCN parameters, i.e., $\mathbf{W}^{(1)} = \mathbf{W}$, it is only necessary to ensure that, with an identity activation $\sigma_I(\cdot)$, $\sigma_I(\mathbf{P_1}\mathbf{Z}\mathbf{W}^{(1)})$ is as close to a zero matrix as possible. Thus, Equation~\eqref{eq_adver_goal_K_1} can be further simplified to:
\begin{equation}\label{eq_adver_goal_0}
   \arg\min_{\mathbf{P}} \mathrm{MSE}( \sigma_I (\mathbf{P_1}\mathbf{Z}\mathbf{W}^{(1)}),[0]_N).
\end{equation}
Upon observation, we consider $\mathbf{P_1}$ can represent a distinct graph $\mathcal{G}_\mathbf{P_1}$, which has the same number of nodes as $\mathcal{G}$, with most nodes being isolated. Perturbations can be viewed as edges, where inserted edges (i.e., when an edge is inserted between nodes $v_i$ and $v_j$, $\mathbf{P_1}_{i,j}=1$) have a weight of $1$, and deleted edges (i.e., when an edge is deleted between nodes $v_i$ and $v_j$, $\mathbf{P_1}_{i,j}=-1$) have a weight of $-1$. Therefore, the equation can be approximated as another GCN training paradigm. The difference is that the training parameters have changed from $\mathbf{W}$ to $\mathbf{P_1}$. At this point, since $\mathcal{G}_\mathbf{P_1}$ is a disconnected subgraph of $\mathcal{G}$ with $r$ edges, hence, when $K=1$, the maximum number of disconnected subgraphs of $\mathcal{G}$ with $r$ edges is the maximum number of forward propagations that the attack model needs to compute. That is, $\mathrm{Cost}(\mathbf{A},r,1) \leq C(|\mathcal{V}|,r)$.

\begin{figure}[htb]
\centering                                                                                                                                                                                                                                                                                                                                                                                                                             \includegraphics[width=0.46\textwidth]{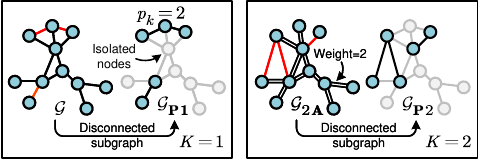}
\caption{For $K=1$ and $K=2$, the primary number of forward propagations for the attacker is determined by $\mathcal{G}_{\mathbf{P_1}}$ and $\mathcal{G}_{\mathbf{P_2}}$, irrespective of the graphs $\mathcal{G}$ and $\mathcal{G}_{2\mathbf{A}}$ that constitute $\mathcal{G}_{\mathbf{P_1}}$ and $\mathcal{G}_{\mathbf{P_2}}$.}
\label{fig_GG2AGP}
\end{figure}

Proceeding to the case where $K=2$, with $\mathbf{P_2}$ which plays the same role of $\mathbf{P_1}$, Equation~\eqref{eq_adver_goal_simply} can be reformulated as follows:
\begin{equation}\label{eq_adver_goal_k_2}
  \arg\min_{\mathbf{P_2}} \mathrm{MSE}(\sigma(\mathbf{A}^2 \mathbf{Z} \mathbf{W}_2  + \mathbf{A}\mathbf{P_2} \mathbf{Z} \mathbf{W}_2 + \mathbf{P_2}^2 \mathbf{Z} \mathbf{W}_2 ),\mathbf{Y}').
\end{equation}
Analogous to the scenario where $K=1$, the objective of the adversary for $K=2$ can be simplified to:
\begin{equation}\label{eq_adver_goal_K_2_0}
  \arg\min_{\mathbf{P_2}} \mathrm{MSE}(\sigma_I((2\mathbf{A} + \mathbf{P_2} )\mathbf{P_2} \mathbf{Z} \mathbf{W}_2 ),[0]_N).
\end{equation}
This expression can be interpreted as an extension of the $K=1$, case: an attack on a graph $\mathcal{G}_{2\mathbf{A}}$ with an adjacency matrix as $2\mathbf{A}$ The maximum number of forward propagations for the attacker is equal to the number of largest disconnected subgraphs in $\mathcal{G}_{2\mathbf{A}}$, which is the same as the $K=1$ case. We present illustrative examples of $\mathcal{G}$, $\mathcal{G}_{\mathbf{P_1}}$, $\mathcal{G}_{\mathbf{P_2}}$ and $\mathcal{G}_{2\mathbf{A}}$ in Figure~\ref{fig_GG2AGP}.

 \begin{figure*}[htp]
    \centering
     \subfigure[Values of $\mathbf{d}_{k,k_{gap}}$ under different $k$ and $k_{gap}$ settings.]{
        \includegraphics[width=0.69\textwidth]{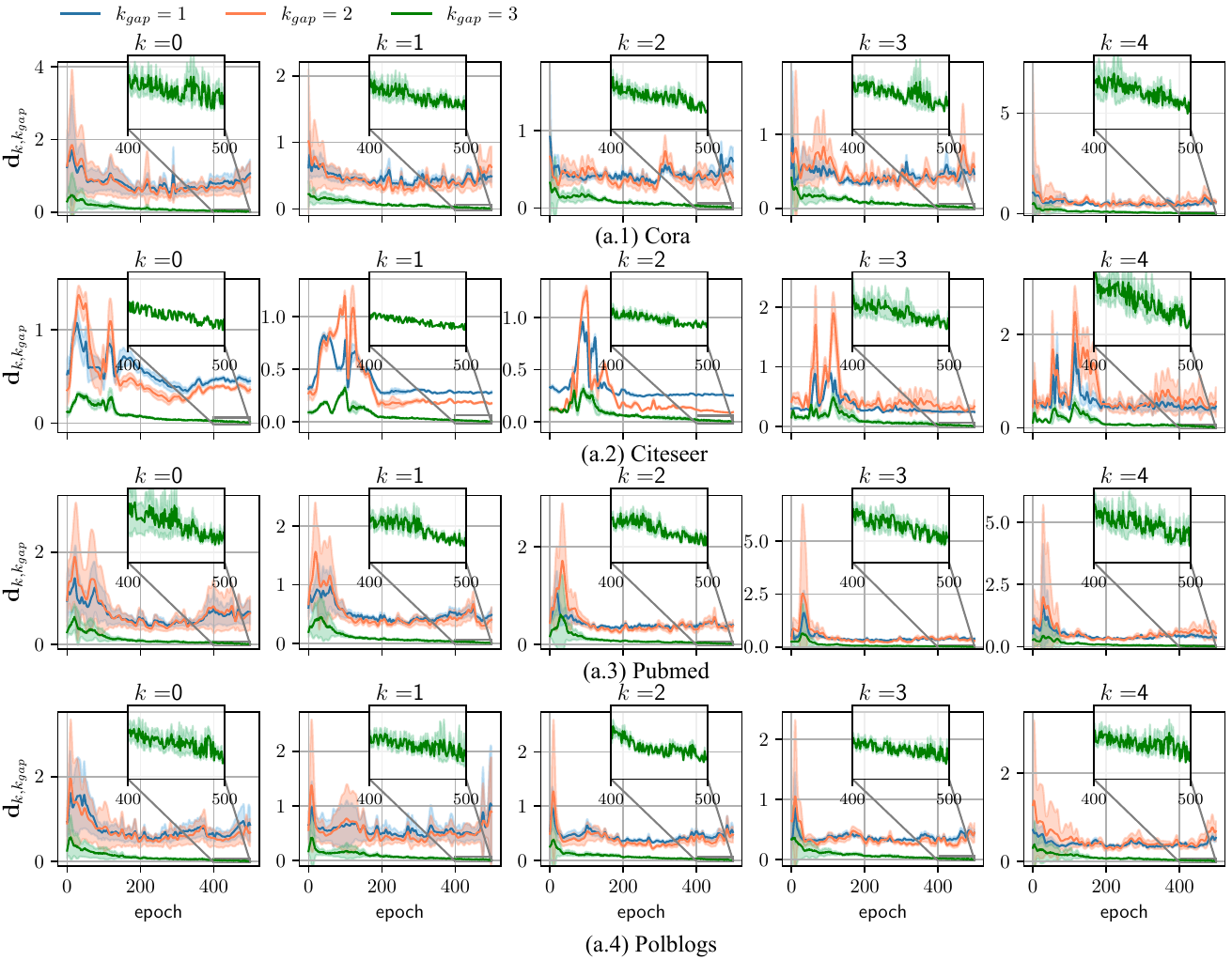}
        \label{fig_add_exp_thm2}
    }%
     \subfigure[Strength distribution of $\mathcal{G}$, $\mathcal{G}_{LRS}$ and $\mathcal{G}_{random}$.]{
        \includegraphics[width=0.29\textwidth]{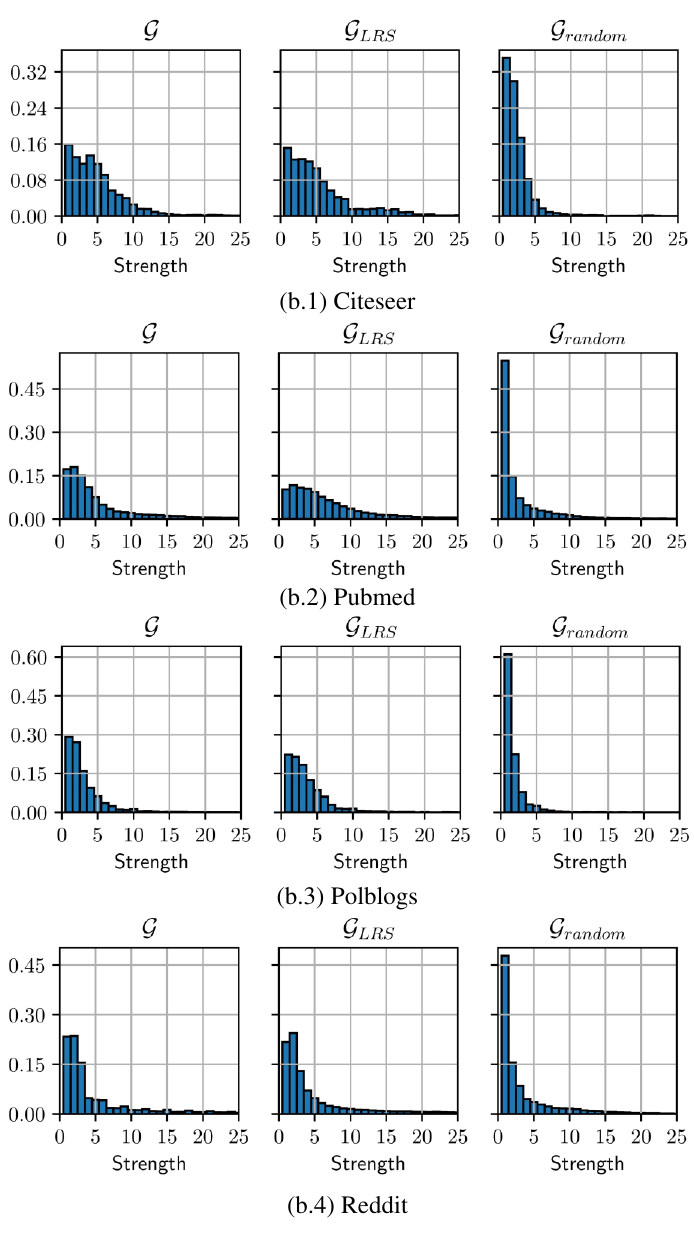}
        \label{fig_strength_add}
    }%
    \caption{Additional Experimental results.}
    \label{fig_add_exp}
\end{figure*}

The situation differs when $K=3$. Specifically, Equation~\eqref{eq_adver_goal_simply} for $K=3$ is given by:
\begin{multline}\label{eq_adver_goal_k_3}
  \arg\min_{\mathbf{P_3}} \mathrm{MSE}(\sigma(\mathbf{A}^3 \mathbf{Z} \mathbf{W}_3  + \mathbf{A}^2\mathbf{P_3} \mathbf{Z} \mathbf{W}_3 \\
  + \mathbf{A}\mathbf{P_3}^2 \mathbf{Z} \mathbf{W}_3 + \mathbf{P_3}^3 \mathbf{Z} \mathbf{W}_3 ),\mathbf{Y}').
\end{multline}
In the above expression, it is required to ensure:
{\small
\begin{equation}\label{eq_adver_goal_K_3_0}
\arg\min_{\mathbf{P}} \mathrm{MSE}( \sigma_I (\mathbf{A}^2\mathbf{P} \mathbf{Z} \mathbf{W}_3 + \mathbf{A}\mathbf{P}^2 \mathbf{Z} \mathbf{W}_3 + \mathbf{P}^3 \mathbf{Z} \mathbf{W}_3),[0]_N).
\end{equation}
}
It can be observed that for $K=3$ the maximum number of connected subgraphs in $\mathcal{G}_{\mathbf{P_3}}$ depends on the number of edges in the graph $\mathcal{G}_{\mathbf{A}^2}$, represented by $\mathbf{A}^2$, rather than the number of edges in $\mathcal{A}$ (which is the case for $K=1$ and $K=2$). In $\mathbf{A}^2$, each element $\mathbf{A}^2_{i,j}$ represents the number of paths of length 2 from node $v_i$ to $v_j$ in $\mathcal{G}_{\mathbf{A}^2}$. Therefore, the number of non-zero elements in $\mathbf{A}^2$ can be at most $\mathcal{V}^2$, as each edge can form a path of length 2 with the other $\mathcal{V}-1$ nodes. Hence, the upper bound on the number of edges contained in $\mathcal{G}_{\mathbf{A}^2}$ is $\frac{|\mathcal{V}|^2}{2}$. The maximum number of disconnected $r$-edges subgraphs (i.e., $\mathcal{G}_{\mathbf{P_3}}$) is $C(\frac{|\mathcal{V}|^2}{2},r)$.

Upon obtaining $\mathcal{G}_{\mathbf{P_3}}$, it needs to be sent to the last two terms, namely $\mathbf{A}\mathbf{P}^2 \mathbf{Z} \mathbf{W}_3$ and $\mathbf{P}^3 \mathbf{Z} \mathbf{W}_3$ for computation. Considering that $\mathbf{P_3}$ is a sparse matrix, $\mathbf{P_3}^3$ is generally a zero matrix, implying that all nodes in the graph $\mathcal{G}_{\mathbf{P_3}^3}$ formed by it are isolated. Consequently, the last two terms can be reduced to one terms, meaning that after obtaining $\mathcal{G}_{\mathbf{P_3}}$, one more matrix multiplication-based forward propagations are needed. In summary, $\mathrm{Cost}(\mathbf{A},r,3) \leq 2C\left(\frac{|\mathcal{V}|^2}{2},r\right)$.

Utilizing the same methodology, for $K>3$, each decomposition introduces a graph $\mathcal{G}_{\mathbf{A}^{K-1}}$ with $\mathbf{A}^{K-1}$ as the adjacency matrix, and a linearly increasing number of summation terms. Combining the above analysis, we can conclude that the maximum number of edges in $\mathcal{G}_{\mathbf{A}^{K-1}}$, grows exponentially, and the number of matrix multiplication-based forward propagation summation terms determined by the maximum number of disconnected subgraphs $\mathcal{G}_{\mathbf{P_r}}$ in $\mathcal{G}_{\mathbf{A}^{K-1}}$ grows linearly. That is, when $K \geq 3$, $\mathrm{Cost}(\mathbf{A},r,K)\leq (K-1) C\left(\frac{|\mathcal{V}|^{K-1}}{2},r\right)$, which completes the proof.

\subsubsection{Proof of Proposition~\ref{pro_E_overhead}}
We adopt an intuitive approach to simplify this computation. Given that $\mathcal{G}_{j,k}$ contains one fewer edge than $\mathcal{G}$ (i.e., $(v_j,v_k)$), $\mathbf{A}_{\mathcal{G}^{j,k}}$ and $\mathbf{A}$ differ in only two elements, hence the subtraction $\mathbf{A}-\mathbf{A}_{\mathcal{G}^{j,k}}$ results in a sparse matrix $\mathbf{D}^{j,k}$, where the majority of elements are zero except for the elements at the $j$-th row and $k$-th column, and $k$-th row and $j$-th column, which are 1.
Given that the parameters of $\mathbf{W}^{(\ell-1)}$ have been trained during the calculation of the $\ell$-th layer's edge-transmitted signals, we use $\mathbf{Z}_{\mathbf{W}}^{(\ell-1)} = \mathbf{Z}^{(\ell-1)}\mathbf{W}^{(\ell-1)}$ to represent the constant matrix in the perspective of the $\ell$-th layer. Then, Equation~\eqref{eq_Z_G_jk} simplifies to
\begin{equation}\label{eq_Z_G_jk_sim}
\mathbf{Z}^{(\ell)}_{\mathcal{G}^{j,k}}=\mathbf{A}\mathbf{Z}_{\mathbf{W}}^{(\ell-1)} - \mathbf{D}^{j,k}\mathbf{Z}_{\mathbf{W}}^{(\ell-1)}.
\end{equation}
Given that $\mathbf{D}^{j,k}$ is a symmetric matrix with only two non-zero values, the result of $\mathbf{D}^{j,k}\mathbf{Z}_{\mathbf{W}}^{(\ell-1)}$, defined as
\begin{equation}\label{eq_R_ijmn}\small
  Q(\mathbf{Z}{\mathbf{W}}^{(\ell-1)};j,k)=[\mathbf{0},\ldots,\underbrace{[\mathbf{Z}_{\mathbf{W}}^{(\ell-1)}]_{k,:} }_{j\text{-th row}},\ldots,\underbrace{[\mathbf{Z}_{\mathbf{W}}^{(\ell-1)}]_{j,:} }_{k\text{-th row}},\ldots,\mathbf{0}]^{\top},
\end{equation}
does not need to be recalculated for each $v_j$, $v_k$, but can instead be reassembled by indexing the rows of $\mathbf{Z}_{\mathbf{W}}^{(\ell-1)}$ according to the chosen of $v_j$, $v_k$, i.e.,

Through the method stated above, Equation~\eqref{eq_Z_G_jk} is eventually simplified to the conclusion of the Proposition~\ref{pro_E_overhead} (Equation~\eqref{eq_E_overhead}).
It can be observed that Equation~\eqref{eq_E_overhead} encompasses all arbitrary selections of $v_j$ and $v_k$, but only the $\mathbf{Z}_{\mathbf{W}}^{(\ell-1)}$ and $\mathbf{A}_{G}\mathbf{Z}_{\mathbf{W}}^{(\ell-1)}$ term is required to be computed. Thus, during the computation of edge-transmitted signals in GRN, we can reduce the initial $|\mathcal{V}|$ computations down to a single computation. Thereafter, when transitioning among different edges, it is only necessary to conduct a column-wise concatenation and substraction of the matrices.

\subsection{Additional Experiments}

In this section, we elucidate the behavior of layers arising from resonance across an expanded range of datasets. Concurrently, we delineate the strength distribution of LRS-constructed graphs drawn from additional datasets. The experimental framework employed here remains consistent with the main body of the text. It is imperative to note that, in the interest of maintaining methodological rigor and fairness, these experiments were conducted based on randomized initializations that have been reset. Consequently, under identical experimental conditions, there exists a slight discrepancy between these results and those presented in the primary narrative. Figures~\ref{fig_add_exp_thm2} and~\ref{fig_strength_add}, which can be conceptually viewed as extensions of Figures~\ref{fig_exp_thm2_main} and~\ref{fig_LRS_strength} respectively, depict these outcomes. The conclusions derived from the illustrated results demonstrate a congruence with the central assertions delineated in the main discourse.

\section*{References}

\begin{itemize}\small
  \item[\lbrack 1\rbrack] Breuer Heinz-Peter, Huber Wolfgang, and Petruccione Francesco. 1994. Fluctuation effects on wave propagation in a reaction-diffusion process. \emph{Physica D}, 73(3): 259--273.
  \item[\lbrack 2\rbrack] Golub Gene H and Reinsch Christian. 1971. Singular value decomposition and least squares solutions. \emph{Linear algebra}.
  \item[\lbrack 3\rbrack] Vannieuwenhoven Nick, Vandebril Raf and Meerbergen Karl. 2012. A new truncation strategy for the higher-order singular value decomposition. \emph{SIAM J. Sci. Comput.}, 34(2): 1027--1052.
  \item[\lbrack 4\rbrack] Al-Mohy, Awad H and Higham, Nicholas J. 2011. Computing the action of the matrix exponential, with an application to exponential integrators. \emph{SIAM J. Sci. Comput.}, 33(2): 488--511.
  \item[\lbrack 5\rbrack] Sun Lichao, Dou Yingtong, Yang Carl, \emph{et al}. 2022. Adversarial attack and defense on graph data: A survey. \emph{IEEE Trans. Knowl. Data Eng.}, 35(8): 7693--7711.
  \item[\lbrack 6\rbrack] Z{\"u}gner Daniel, Akbarnejad Amir and G{\"u}nnemann Stephan. 2018. Adversarial attacks on neural networks for graph data. \emph{Proc. 24th ACM SIGKDD Int. Conf. Knowl. Discov. Data Min.}, 2847--2856.
\end{itemize}

\end{document}